\pdfoutput=1
\documentclass[11pt]{article}
\usepackage{amsmath,amsfonts,bm}

\def\eqref#1{equation~\ref{#1}}
\def\1{\bm{1}}

\DeclareMathAlphabet{\mathsfit}{\encodingdefault}{\sfdefault}{m}{sl}
\SetMathAlphabet{\mathsfit}{bold}{\encodingdefault}{\sfdefault}{bx}{n}

\usepackage{ACL2023}

\usepackage{times}
\usepackage{latexsym}
\usepackage{xr}

\usepackage[T1]{fontenc}
\usepackage[utf8]{inputenc}

\usepackage{microtype}

\usepackage{inconsolata}

\usepackage{subfiles}

\usepackage{caption}
\usepackage{subcaption}
\usepackage{enumitem}
\usepackage{flushend}
\usepackage{balance}
\usepackage{lineno}
\usepackage{amsmath,amssymb,amsfonts}
\usepackage{algorithm}
\usepackage{algorithmic}
\usepackage{graphicx}
\usepackage{textcomp}
\usepackage{xcolor}
\usepackage{colortbl}
\usepackage{amsthm}
\usepackage{url}
\usepackage{float}
\usepackage{multirow}
\usepackage{multicol}
\usepackage{color}
\usepackage{bm}
\usepackage{bbm}

\usepackage{pifont}
\usepackage{array}
\newcolumntype{L}[1]{>{\raggedright\let\newline\\\arraybackslash\hspace{0pt}}m{#1}}
\newcolumntype{C}[1]{>{\centering\let\newline  \\\arraybackslash\hspace{0pt}}m{#1}}
\newcolumntype{R}[1]{>{\raggedleft\let\newline \\\arraybackslash\hspace{0pt}}m{#1}}
\usepackage{ bbold }

\usepackage[utf8]{inputenc} % allow utf-8 input
\usepackage[T1]{fontenc}    % use 8-bit T1 fonts
\usepackage{hyperref}       % hyperlinks
\hypersetup{colorlinks,linkcolor={red},urlcolor={blue}} 
\usepackage{url}            % simple URL typesetting
\usepackage{booktabs}       % professional-quality tables
\usepackage{amsfonts}       % blackboard math symbols
\usepackage{nicefrac}       % compact symbols for 1/2, etc.
\usepackage{microtype}      % microtypography
\usepackage{xcolor}         % colors
\usepackage{xr}

\usepackage{pdfpages}
\usepackage{amssymb}

\usepackage[flushleft]{threeparttable}
\definecolor{Gray}{gray}{0.9}

\title{Large Language Models for Data Annotation and Synthesis: A Survey}

\author{
  Zhen Tan\textsuperscript{\ding{171}}\thanks{\ \ Equal contribution.} ,  
  Dawei Li\textsuperscript{\ding{171}}\footnotemark[1] ,   
  Song Wang\textsuperscript{\ding{168}}\footnotemark[1] ,  
  Alimohammad Beigi\textsuperscript{\ding{171}} , 
  \textbf{Bohan Jiang}\textsuperscript{\ding{171}} ,\\
  \textbf{Amrita Bhattacharjee}\textsuperscript{\ding{171}}, 
  \textbf{Mansooreh Karami}\textsuperscript{\ding{171}}, 
  \textbf{Jundong Li}\textsuperscript{\ding{168}}, 
  \textbf{Lu Cheng}\textsuperscript{\ding{170}}, 
  \textbf{Huan Liu}\textsuperscript{\ding{171}} \\
  \textsuperscript{\ding{171}}School of Computing, and Augmented Intelligence, Arizona State University\\
  \textsuperscript{\ding{168}}Department of Electrical and Computer Engineering, the University of Virginia \\
  \textsuperscript{\ding{170}}Department of Computer Science, University of Illinois Chicago\\
  {\tt \{ztan36,abeigi,abhatt43,bjiang14,mkarami,huanliu\}@asu.edu}\\
  {\tt \{sw3wv,jundong\}@virginia.edu}
  {\tt lucheng@uic.edu}\\
}

\begin{document}

\maketitle

\begin{abstract}
Data annotation and synthesis generally refers to the labeling or generating of raw data with relevant information, which could be used for improving the efficacy of machine learning models. The process, however, is labor-intensive and costly. The emergence of advanced Large Language Models (LLMs), exemplified by GPT-4, presents an unprecedented opportunity to automate the complicated process of data annotation and synthesis. While existing surveys have extensively covered LLM architecture, training, and general applications, we uniquely focus on their specific utility for data annotation. This survey contributes to three core aspects: LLM-Based Annotation Generation, LLM-Generated Annotations Assessment, and LLM-Generated Annotations Utilization. Furthermore, this survey includes an in-depth taxonomy of data types that LLMs can annotate, a comprehensive review of learning strategies for models utilizing LLM-generated annotations, and a detailed discussion of the primary challenges and limitations associated with using LLMs for data annotation and synthesis. Serving as a key guide, this survey aims to assist researchers and practitioners in exploring the potential of the latest LLMs for data annotation, thereby fostering future advancements in this critical field\footnote{We release the paper list of LLM-based data annotation and synthesis at \url{https://github.com/Zhen-Tan-dmml/LLM4Annotation}}.  
\end{abstract}

% \documentclass[../LLM4Annotate.tex]{subfiles}

% \begin{document}

\section{Introduction}

%1. What is data annotation (list of annotaion)
%2. Why data annotation is hard for current machine learning models
% 3. LLM is powerful and can be used for DA

In the complex realm of machine learning and natural language processing (NLP), data annotation and synthesis stand out as a critical yet challenging task, extending beyond simple label attachment to encompass a diverse array of fundamental or auxiliary information. This detailed process typically involves \ding{182} categorizing raw data with class or task labels for basic classification, \ding{183} adding intermediate labels for contextual depth~\cite{yu2022generate}, \ding{184} assigning confidence scores to assess annotation reliability~\cite{lin2022teaching}, \ding{185} applying alignment or preference labels to tailor outputs to specific criteria or user needs, \ding{186} annotating entity relationships to understand how entities within a dataset interact with each other~\cite{wadhwa2023revisiting}, \ding{187} marking semantic roles to define the underlying roles that entities play in a sentence~\cite{larionov2019semantic}, \ding{188} tagging temporal sequences to capture the order of events or actions~\cite{yu2023temporal}, or \ding{189} Synthesize data in the format of instruction~\cite{wang2022self}, response~\cite{zhang2023self}, reasoning~\cite{wang2022pinto}, pairwise~\cite{bai2022constitutional} and textual feedback~\cite{pan2024automatically} to for language model tuning.

\begin{figure*}[!t]
\vspace{-2mm}
	\centering
	\includegraphics[width=0.99\linewidth]{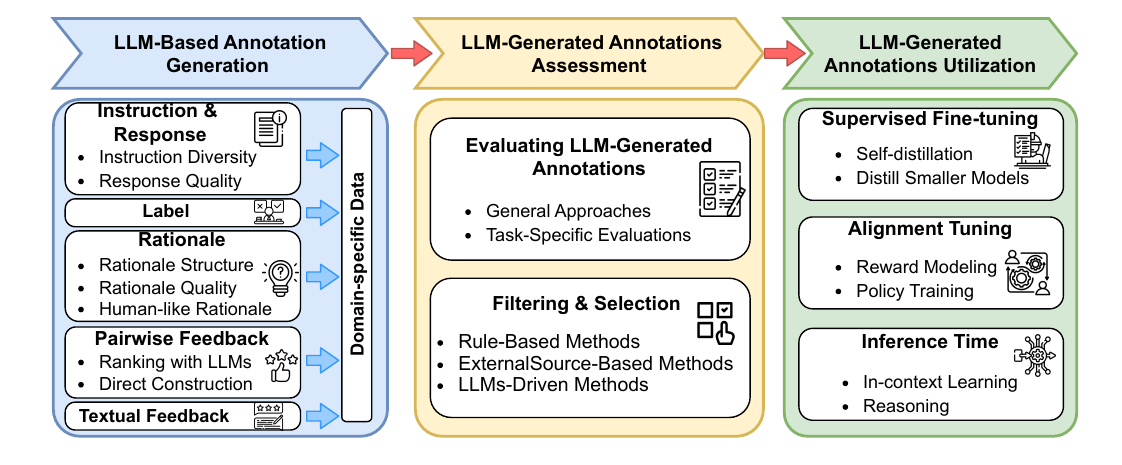}
    \vspace{-2mm}
	\caption{The proposed taxonomy of existing research on LLM for data annotation.}
	\label{fig:intro}
	\vspace{-5mm}
\end{figure*}

 Despite its wide applications, data annotation and synthesis poses significant challenges for current machine learning models due to the complexity, subjectivity, and diversity of data~\cite{yang2023new}. This process requires domain expertise and is resource-intensive, particularly when manually labeling or creating large datasets. Advanced LLMs such as GPT-4~\cite{openai2023gpt4}, Gemini~\cite{team2023gemini}, and LLaMA-2~\cite{touvron2023llama2} offer a promising opportunity to revolutionize data annotation. LLMs serve as more than just tools but play a crucial role in improving the effectiveness and precision of data annotation. Their ability to automate annotation tasks, ensure consistency across large volumes of data~\citep{hou2023large}, and adapt through fine-tuning or prompting for specific domains~\citep{song2023preference,Zhang2024BalancingSA}, significantly mitigates the challenges encountered with traditional annotation and synthesis methods, setting a new standard for what is achievable in the realm of NLP.
%add citation 
This survey delves into the nuances of using LLMs for data annotation and synthesis, exploring methodologies, utilizing strategies, and associated challenges in this transformative approach. Through this exploration, we aim to shed light on the motivations behind embracing LLMs as catalysts for redefining the landscape of data annotation and synthesis in machine learning and NLP.
We explore the utilization of LLMs for annotation synthesis in this survey, making four main contributions:
\vspace{-3mm}
\begin{itemize}[leftmargin=*]
    \item \textbf{LLM-Based Annotation Generation:} We dive into the process of synthesizing annotations for various data types, including instruction \& response, rationale, pairwise feedback, textual feedback, and other domain-specific data. Additionally, we discuss the criteria (\emph{e.g.,} diversity and quality) in the annotation process.
    \vspace{-3mm}
    \item \textbf{Assessing LLM-Generated Annotations:} We explore various methods for assessing the quality of annotations and strategies for selecting high-quality annotations from numerous options.
    \vspace{-3mm}
    \item \textbf{LLM-Generated Annotations Utilization:} We investigate the methodologies at different stages, including supervised fine-tuning, alignment tuning, and inference time, to train machine learning models based on LLM-generated annotations.
    \vspace{-3mm}
    \item \textbf{Social Impact and Future Work:} We discuss issues ranging from ethical dilemmas, such as bias and implications, to technical limitations, including hallucination and efficiency in LLM-generated annotations.
\end{itemize}
\vspace{-3mm}
Focusing on this underrepresented aspect of LLM application, the survey aims to serve as a valuable guide for academics and practitioners who intend to deploy LLMs for annotation purposes. Note that in this survey, we primarily focus on pure language models and do not extensively cover recently emerging multimodal LLMs, such as LLaVA~\cite{liu2023visual}. Figure~\ref{fig:intro} illustrates the general structure of this survey. Additionally, a list of potential tools for utilizing LLMs for annotation is included in Appendix~\ref{app:tool}, along with explanatory examples.

\noindent\textbf{Differences from Other LLM-related Surveys.} While existing surveys in the NLP domain extensively cover architectural nuances~\cite{zhao2023survey}, training methodologies~\cite{liu2023trustworthy}, and evaluation protocols~\cite{chang2023survey} associated with LLMs, their main focus lies on the capabilities of models for specific end tasks such as machine translation~\cite{min2021recent}, alignment~\cite{wang2023aligning}, code generation~\cite{zan2023large}, and medical analysis~\cite{thirunavukarasu2023large}. In contrast, this survey distinguishes itself by focusing primarily on the application of these potent next-generation LLMs to the intricate realm of annotation synthesis, a domain that is crucial yet underexplored.
\vspace{-2mm}
% Include the PDF

% \end{document}

% \documentclass[../LLM4Annotate.tex]{subfiles}

% \begin{document}

\section{Preliminaries}\label{sec:formulation}
% \vspace{-2mm}
% In this section, we introduce significant notations utilized in this paper and preliminaries. The notations and their definitions can be found in Table~\ref{tb:symbols}.
% \vspace{-6mm}

% \subsection{Problem framework}
In this section, we delve into our approach to the annotation synthesis process. We introduce two core models: an annotator model, denoted as $\mathcal{A}$, which maps input data to annotations, and a task learner, represented as $\mathcal{L}$, that utilizes or learns from these annotated data to accomplish specific tasks. Our primary focus is on utilizing advanced LLMs like GPT-4~\citep{openai2023gpt4} and LLaMA~\citep{touvron2023llama} as annotators ($\mathcal{A}$), while the task learner ($\mathcal{L}$) can be another large model~\cite{vicuna2023} or a less complex one such as BERT~\citep{devlin2018bert}, which utilizes these annotated data to perform designated tasks. LLM-generated annotations encompass categorical labels and enhance raw data points with a comprehensive array of auxiliary signals. These annotations, including confidence scores, contextual details, and other metadata, extend beyond traditional categorical labels.

\vspace{-2mm}
\section{LLM-Based Annotation Generation}
\vspace{-2mm}
The emergence of LLMs has sparked significant interest in their capacity for high-quality, context-sensitive annotation synthesis. This section discusses various kinds of annotations and data produced via LLMs.

\subsection{Instruction \& Response}

Instruction and response are the two fundamental components that constitute a dataset for LLM fine-tuning and in-context learning (ICL).
Previous NLP datasets~\cite{li2017dailydialog,wang2018glue,ouyang2022training} mainly rely on human annotators to construct.
Recently, with the advent of LLMs, automatic and generative methods~\cite{meng2022generating,ye2022zerogen,ye2022progen,wang2024codeclm,wu2024unigen,liu2024best} have gained more focus in data annotation.
% \lu{The logic does not look right to me. How is LLM data annotation related to data augmentation and generation?}
% \dl{Rvised. Remove discussion about previous data augmentation techniques.}

\noindent\textbf{Instruction Diversity.}
The diversity of instruction has been proven crucial for LLM learning~\cite{li2023self,song2024preference,song2024scaling,tangmathscale}.
Recent studies have explored various methods to diversify and augment instructions in the original datasets.
For example, \citet{yoo2021gpt3mix} enhance data diversity by mixing two different samples to create a new one.
\citet{wang2022self} use a few manually-written seed instructions and iteratively augment them with a generate-then-filter pipeline.
Additionally, \citet{meng2023tuning,wang2023sass} train an instruction generation model in the original dataset to augment the diversity of instruction.
\citet{gupta2023targen} employ a multi-step prompting method to first generate task descriptions, which are then used as instance seeds to guide LLMs in instruction generation.
To obtain informative and diverse examples,~\citet{wang2023let} propose an explain-then-generate pipeline with LLMs for iterative data synthesis.
Besides, \citet{li2023dail} paraphrase the given sample multiple times to help LLMs understand them from different perspectives.
\citet{koksal2024longform} suggest a clustering-based data selection method to ensure diversity in the initial seed data for augmentation.
Recently, \citet{yu2024large} introduce AttrPrompt as an effective way to balance diversity and cost in LLM-based data annotation.
\citet{xu2024magpie} propose to synthesize high-quality instruction data at scale by extracting it directly from an aligned LLM and present a self-synthesis method for generating large-scale alignment data named Magpie.
To improve the diversity,~\citet{chan2024scaling} introduce Persona Hub -- a collection of 1 billion diverse personas automatically curated from web data, to foster the creation of diverse synthetic data at scale for various scenarios.
\citet{zhu2024fanno} introduce FANNO, a fully autonomous, open-sourced framework that revolutionizes the annotation process without the need for pre-existing annotated data.
Recently,~\citet{kowshik2024corrsynth} propose CorrSynth, which generates data that is more diverse and faithful to the input prompt using a correlated sampling strategy.
To augment the clinical data safely and protect the patient's privacy,~\citet{} propose to use GPT-4 for data augmentation through one-shot and zero-shot prompts.
~\citet{divekar2024synthesizrr} propose Synthesize by Retrieval and Refinement (SynthesizRR), which uses retrieval augmentation to introduce variety into the dataset synthesis process: as retrieved passages vary, the LLM is seeded with different content to generate its examples.
~\citet{li2024optimizinga} introduce a general and scalable framework, IDEA-MCTS (Instruction Data Enhancement using Monte Carlo Tree Search), a scalable framework for efficiently synthesizing instructions.

\noindent\textbf{Response Quality.}
High-quality responses are essential for effective fine-tuning and ICL~\cite{luo2024assessing}.
To improve the quality of the generated response,~\citet{zhang2023self} frame the response generation as reading comprehension tasks and create detailed prompts for LLMs.
\citet{huang2023large} adopt self-consistency~\cite{wang2022self} in response generation, selecting from the candidate response with the highest confidence score.
Furthermore, \citet{yang2024self} propose self-distill and augment the instruction tuning dataset by rewriting the original responses. 
\citet{pang2024self} conduct social simulations to ensure high-quality, human-valued responses from LLMs.
Moreover, \citet{liu2024mixture} introduce a multi-step prompting including question analysis, answer guidance and safe answer production in their response generation pipeline.
\citet{guo2024human} enhance the LLMs outputs' quality by implementing retrieval-augmented ICL and providing LLMs with relevant documents.
To ensure LLMs provide responses aligned with human values,~\citet{sun2024principle} and~\citet{wang2024step} conduct principle-driven prompting, guiding LLMs with well-crafted and detailed principles.
Besides,~\citet{lupidi2024source2synth} propose Source2Synth, which takes as input a custom data source and produces synthetic data points with intermediate reasoning steps grounded in real-world sources.
Recently,~\citet{chen2024controlmath} follow ``less is more'' principle, achieving better results with fewer data points in mathematical reasoning data generation.
%To evaluate the response quality regarding empathy, \citet{luo2024assessing} incorporate novel empathy ranking evaluation (EMRank) involving automated metrics measured by LLaMA which are shown to reach a desired agreement with human evaluation. 

\subsection{Label}
Label is an important component of the traditional classification task in NLP.
Nowadays, many researchers focus on automating label annotation with the assistance of LLMs~\citet{yadav2024towards,tseng2024expert}.
\citet{chen2024large} introduce an innovative approach where we employ LLMs as expert annotators for event extraction.
\citet{martorana2024zero} propose a method to support metadata enrichment using topic annotations generated by several LLMs.
Both \citet{wu2024enhancing} and \citet{ahmed2024can} explore the potential of large language models (LLMs) as automated data annotators to improve efficiency and consistency in label annotation tasks.
One interesting work from~\citet{li2023coannotating} proposes CoAnnotating, a novel paradigm for Human-LLM co-annotation of unstructured texts at scale.
Moreover,~\citet{tekumalla2023leveraging} evaluate the utilization of LLM in labeling COVID-19 vaccine-related tweets, with the purpose of comparing performance against human annotators.
To address the potential limitation of LLMs' annotation,~\citet{tornberg2024best} propose a comprehensive set of standards and best practices for their reliable, reproducible, and ethical use.
Additionally, there are also some works that utilize LLMs to improve the original annotation made by human annotators~\cite{laskar2023can,flamholz2024large,wang2024model}.
To reduce costs,~\citet{schmidt2024prompting} argue that domain-agnostic knowledge from LMs, such as linguistic understanding, is sufficient to create a well-curated dataset.
Recently,~\citet{choi2024unigen} propose a novel approach to universal domain generalization that generates a dataset regardless of the target domain.
~\citet{li2024optimizing} integrate LLMs with a novel sorting method to address the multi-level function call relationships within repositories for code annotation.
~\citet{choi2024multi} leverage approaches such as chain-of-thought and majority voting to imitate human annotation and classify unrelated documents from the Multi-News dataset, which is widely used for the multi-document summarization task.
To calibrate models and improve their generalization,~\citet{ba2024fill} use LLMs as synthetic data generation strategies to lower the ECE bound and improve model accuracy on real test data.

\subsection{Rationale}
% \lu{Same question here, How is LLM data annotation related to rationale generation?}
% \dl{Revised. Define the rationale generated by LLMs as a kind of auxiliary information.}
The rationale reflects the detailed thought process and reasoning pathway an individual follows when solving a given question, being considered valuable auxiliary information for the final answer prediction.
In early studies~\cite{ling2017program,cobbe2021training,wei2022chain}, the rationale in each dataset was annotated by human experts, significantly limiting its availability and scalability.
\citet{kojima2022large} initially confirm the efficacy of the chain-of-thought (CoT) approach in LLMs and boosting LLMs' reasoning through the integration of self-generated rationales.

\begin{figure*}[!t]
\vspace{-2mm}
	\centering
	\includegraphics[width=0.99\linewidth]{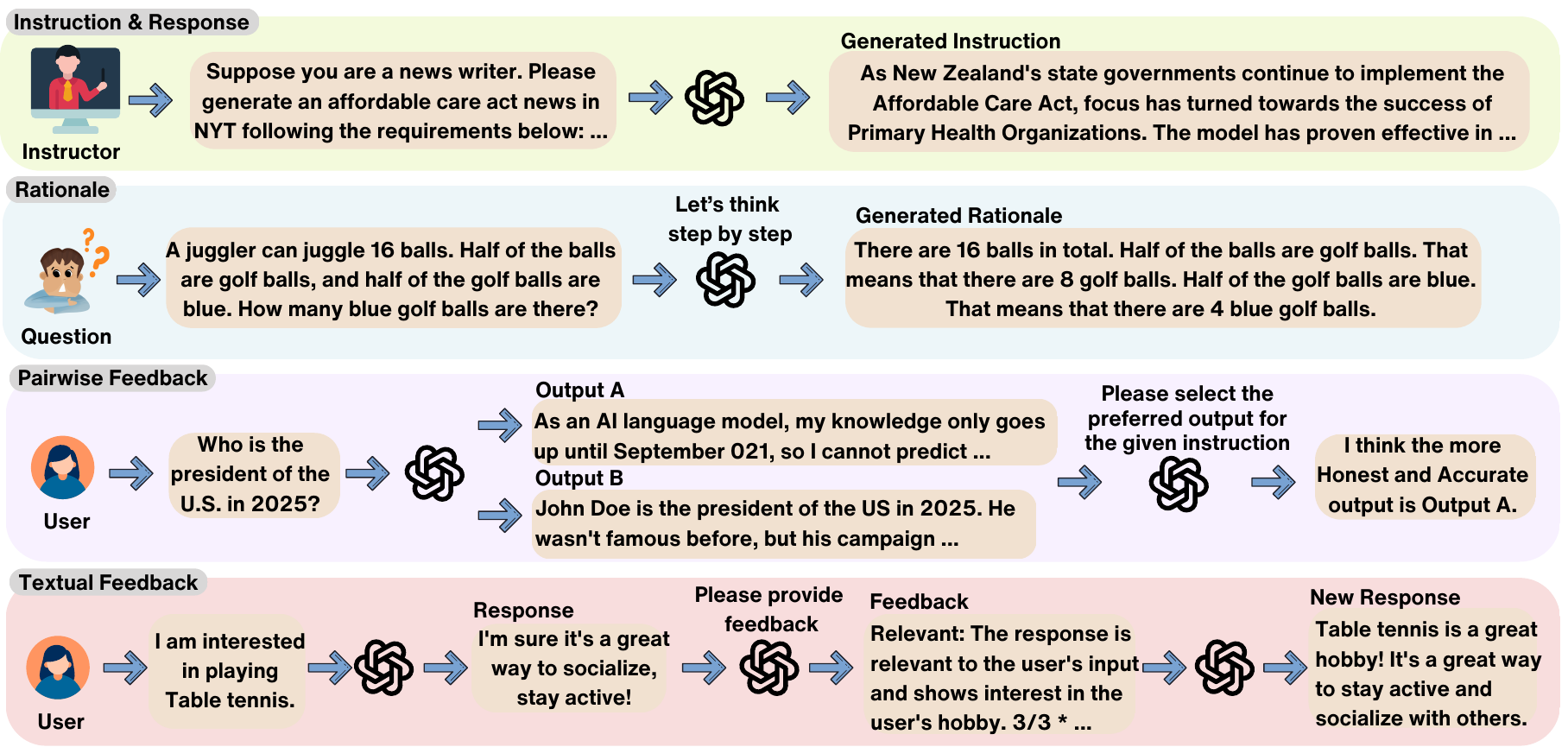}
    \vspace{-2mm}
	\caption{The examples for LLM-based annotation generation.}
	\label{fig:examples}
	\vspace{-5mm}
\end{figure*}
\noindent\textbf{Rationale Structure.} Following~\citet{kojima2022large}, there is a notable interest in abstracting the reasoning process of LLMs into diverse structures and format, including trees~\cite{hao2023reasoning,yao2024tree}, graphs~\cite{besta2024graph,yao2023beyond}, tables~\cite{wang2024chain}, programs~\cite{chen2023program}, recursion~\cite{qi2023art}, 
and concepts~\cite{tan2023interpreting}.

\noindent\textbf{Rationale Quality.} To produce high-quality and fine-grained rationale, diverse methodologies have been employed.
\citet{wang2022pinto} prompt frozen LLMs to produce choice-specific rationales to elucidate each choice in a sample.
\citet{wang2023scott} employ contrastive decoding to foster more plausible rationales, taking into account gold-standard answers.
\citet{liu2023logicot} curate meticulously designed prompts to derive high-quality rationales from GPT-4 and construct a logical CoT instruction tuning dataset.
For attaining fine-grained rationales,~\citet{shridhar2023distilling} introduce Socratic CoT by decomposing the original question into a series of subquestion-solution pairs and generating CoT for them separately.
Additionally,~\citet{kang2024knowledge} propose a neural reranker to acquire supplementary relevant documents for rationale generation in knowledge-intensive reasoning tasks.
Besides,~\citet{zhou2024enhancing} explore the potential and limitations of using graph-based synthetic reasoning data as training signals to enhance LLMs' reasoning capabilities.
Recently,~\citet{wei2024instructrag} propose InstructRAG, where LMs explicitly learn the denoising process through self-synthesized rationales.

\noindent\textbf{Human-like Rationale.} Another intriguing avenue in synthesized rationale delves into making the reasoning process more human-like~\cite{gao2023human}.
Many studies emulate human diverse thinking in problem-solving, sampling multiple reasoning pathways for a given question~\cite{gao2021making,wang2022self,chen2023universal,liu2023plan}.
Subsequent studies~\cite{tong2023eliminating,balepur2023s,ma2023poe} explore the elimination reasoning in LLMs, checking each reasoning pathway reversely and removing the incorrect candidates.
Moreover, various works~\cite{yin2023exchange,liang2023encouraging,xu2023towards,liu2023dynamic} explore the peer collaboration and debate among individual LLMs to capture human-like discussions as rationales.

\subsection{Pairwise Feedback}
While high-quality human feedback is proven to be effective in aligning LLMs' values and preferences with us humans, recent advancements aim to automate this pairwise feedback mechanism.

\noindent\textbf{Ranking with LLMs.} One technique is to sample multiple responses and have the LLM rank these candidates based on various criteria~\cite{bai2022constitutional,lee2023rlaif,yuan2024self}.
\citet{Sun2023SALMONSW} sample two responses from the initial policy model and use the model to select the preferred response based on a human-written principle~\cite{sun2024principle}.
\citet{zhang2024self} propose a self-evaluation mechanism, generating questions for each response and measuring factuality by the LLM's confidence in the answers.
To improve synthetic data quality,~\citet{pace2024west} combine the Best-of-N and Worst-of-N sampling strategies and introduce the West-of-N approach.
They constructed data pairs by identifying the best- and worst-scored responses according to a pre-trained preference model.
In robotics, \citet{zeng2024learning} iteratively update the reward function with the self-ranked responses from LLMs, enhancing learning efficiency without human supervision.

\noindent\textbf{Direct Construction.} Another effort towards automatic pairwise feedback generation involves directly generating responses of various qualities~\cite{feng2024improving,lee2024reinforcement}.
To accomplish this, they typically have to make various assumptions when determining the factors influencing response quality.
For example,~\citet{kim2023aligning} assume larger LLM with more shots will give better responses and produce synthetic pairs based on this.
\citet{tong2024optimizing} follow the rule of thumb that the supervised fine-tuning model will perform better than its unfinetuned base model.
Adhere to this criterion, they start with a few seed data, iteratively training the model and synthesizing comparison data pairs.
\citet{yang2023rlcd} create quality differences by prompting LLMs to either follow or violate given principles.
To measure the response quality more subjectively,~\citet{xu2023contrastive} introduce multiple LLMs and utilize benchmark scores to define superiority.
%\citet{lee2024reinforcement} borrow insights from self-refine and propose reinforcement learning from reflective feedback (RLRF). It utilizes fine-grained feedback, incorporating detailed criteria to enhance responses, and considers the edited versions superior to the original ones.
% \citet{feng2024improving} hold the assumption that (1) the correct given answer more likely leads to correct rationale and (2) the rationale that leads to the correct answer is better than the rationale that leads to the wrong answer, producing comparisons based on these principles.

\subsection{Textual Feedback}
Textual feedback~\cite{pan2024automatically} generated by LLMs typically highlights the shortcomings of the current output or suggests specific improvements, thus offering rich and valuable information for polishing or evaluating the generated response.
Many existing works tailor appropriate prompts and instruct LLMs to generate such informative feedback in various tasks, including question answering~\cite{madaan2024self,shinn2024reflexion}, machine translation~\cite{chen2023iterative,raunak2023leveraging} and hallucination detection~\cite{yang2023new,manakul2023selfcheckgpt}.
Some investigations have explored leveraging debate and peer review as feedback to enhance LLMs' reasoning~\cite{du2023improving,xu2023towards,cohen2023lm,fu2023improving} and evaluation~\cite{li2023prd,chu2024pre,ning2024peer} capabilities.
Additionally, efforts have been made to analyze reasons for undesired or incorrect responses produced by LLMs, thus facilitating reflection and learning from their previous mistakes~\cite{wang2023learning,an2023learning,chen2023gaining,tong2024can}.

\subsection{Other Domain-Specific Data}
Distilling multi-round conversations from LLMs presents a highly cost-effective approach for constructing high-quality dialogue datasets~\cite{kim2023soda,xu2023baize,chen2023places,li2024camel,wang2024ceb,liang2024synth} or enhancing existing ones~\cite{zheng2023augesc,chen2022weakly,zhou2022reflect,Sun2024FosteringNC}.
In graph and tabular data, several studies prompt LLMs to contextualize these structural data~\cite{xiang2022asdot,kim2023soda,li2024contextualization,ronzano2024towards,xiong2023tilp,xiong2024teilp} or distill structural insights from raw text~\cite{bi2024codekgc,li2024dalk,ding2024automated,xiong2024large,tuozzo2022moving}.
Moreover, LLMs have also been widely adopted in the research of robotics and agents, serving as proficient data annotators to generate plans~\cite{huang2022language,brohan2023can,rana2023sayplan,singh2023progprompt,lin2023text2motion}, simulation tasks~\cite{wang2023gensim,ha2023scaling} and supervised signal~\cite{kwon2022reward,du2023guiding}.
Besides, LLMs are acting as efficient data annotators in various artificial intelligence domains, including multi-modal~\cite{li2023stablellava,yin2024lamm,chen2024tomgpt,luo2024mmevol,liu2024synthvlm,wang2024world,cheng2024least}, recommendation system~\cite{acharya2023llm,shen2024pmg,wei2024llmrec,zhang2024large}, information extraction~\cite{josifoski2023exploiting,jeronymo2023inpars,li2024read,ma2024star,bonn2024adjudicating}, multilingual annotation~\cite{frei2023annotated,hamerlik2024chatgpt,latouche2024zero} and etc~\cite{chu2024causal,bhattacharjee2024zero,martorana2024text,zhaoself}.

\vspace{-2mm}
\section{LLM-Generated Annotations Assessment}
\vspace{-1mm}
Effective evaluation of annotations generated by LLMs is crucial to fully harness their potential. This section focuses on two main aspects: 
\vspace{-2mm}
\subsection{Evaluating LLM-Generated Annotations}
\vspace{-1mm}
This subsection explores various methods for assessing annotation quality, ranging from human-led to automated approaches.

\noindent\textbf{General Approaches:} Research has investigated diverse methods for evaluating LLM annotations. The ``Turking Test'' by \citet{efrat2020turking}, evaluates LLMs' adherence to data annotation guidelines, with human annotators comparing LLM outputs against benchmarks like SNLI~\cite{bowman2015large}, SQuAD~\cite{rajpurkar2016squad}, and NewsQA~\cite{trischler2016newsqa}. Similarly, \citet{honovich2022unnatural} manually examined the originality, accuracy, and variety of datasets created by LLMs, focusing on their response to instructions. Additionally, studies such as by \citet{alizadeh2023open} measure the performance of open-source LLMs against human-annotated labels in tasks like relevance and topic detection.

\noindent\textbf{Task-Specific Evaluations:} Methodologies vary by application. For instance, in knowledge graph enhancement, token ranking metrics assess LLM contributions in fact completion. Additionally, evaluations of counterfactual generation often utilize diversity metrics like Self-BLEU \cite{chen2023disco}, while code generation relies on metrics such as Pass@k \cite{nijkamp2022codegen}. In scenarios requiring extensive datasets, the quality of LLM-generated annotations is compared to gold standard labels within a small, labeled subset \cite{zhao2021lmturk, agrawal2022large, he2023annollm}.

\noindent\textbf{LLM-as-a-Judge:}
LLM-as-a-judge~\cite{wu2024meta,zheng2023judging,li2024generation} is a commonly used method in automatic generation evaluation.
To scale the assessment of the synthetic data or annotation, there are also some works that adopt LLM-as-a-judge to conduct the evaluation.
\cite{li2024coevol} employ multiple LLMs to debate with each other to evaluate the synthetic data's quality fairly, iteratively improving response quality, while creating a judge LLM to select preferred responses for enhanced instruction tuning.
To enhance the quality of the synthetic instruction tuning data,~\citet{liang2024sheep} introduce an iterative self-enhancement paradigm (I-SHEEP).
During training, they adopt LLM-as-a-judge to score the synthetic responses and set a threshold to collect high-quality query-response pairs for the subsequent training iteration.

% \vspace{-2mm}
% \subsection{Data Selection via Active Learning}
% Choosing high-quality annotations from numerous options is crucial. Active Learning (AL) emerges as a key technique, especially when integrating LLMs into the AL process. This section introduces pool-based AL within the Learning for Annotation framework, where a vast pool of unlabeled data and a smaller set of labeled data exist. AL strategically selects the most informative samples from the pool to enhance the learning model's performance or until reaching a budget limit.

% \noindent\textbf{LLMs as Acquisition Functions:} Various types of acquisition functions $\alpha(x_i,\gL)$ exist, categorized as (a) Diversity, (b) Uncertainty, and (c) Similarity. Notable research in this context includes studies by \citet{shelmanov2021active, tamkin2022active, margatina2023active}, each investigating different aspects of using LLMs as acquisition functions.

% \noindent\textbf{LLMs as Oracle Annotators:} Innovative studies \cite{bansal2023large, wu2023scattershot} have employed LLMs as oracle annotators in AL setups, enhancing domain generalization and in-context learning for NLP models. Additionally, \citet{kim2023prefer} proposed utilizing LLMs to annotate task-specific preferences between input text pairs, facilitating joint learning with task labels.

\vspace{-2mm}
\subsection{Filtering \& Selection}
Selecting high-quality annotations from numerous options is crucial.
In this section, we categorize the filtering and selection methods for LLM-generated data into three types: rule-based filtering, external source utilization, and LLMs-driven selection.

\noindent\textbf{Rule-Based Methods.}
Rule-based methods follow various heuristic assumptions concerning sample length~\cite{li2023stablellava,kim2023soda}, keyword occurrence~\cite{kim2023aligning,zheng2023augesc} and specific patterns~\cite{zhang2023self,guo2024human,ding2024automated} to filter low-quality or undesiered synthetic data points.
\citet{zheng2023augesc,kim2023soda} establish thresholds for the number of rounds in generated conversations to guarantee each synthetic dialogue is informative enough.
\citet{ho2023large,kang2024knowledge} employ ground truth parsing to filter out incorrect CoT rationales within each candidate reasoning sample.
To encourage diversity among the generated data points,~\citet{wang2022self,lee2023making,ding2024automated} utilize semantic similarity metrics to identify and remove redundant samples.

\noindent\textbf{External-Source-Based Methods.}
There are also many works that depend on the external source's feedback to clean and refine synthetic datasets~\cite{kim2023soda}.
With a pre-trained reward model,~\citet{gulcehre2023reinforced,dong2023raft} augment the original dataset only with samples that obtain high reward values.
When distilling smaller models,~\citet{lin2023selective,wang2024codeclm} meticulously select appropriate data through the feedback from the student models.
Other approaches~\cite{chen2023disco,zheng2023augesc} utilize pre-trained classification models to discern between target and unwanted data points.

\noindent\textbf{LLMs-Driven Methods.}
The versatility of LLMs has invoked interest in leveraging LLMs themselves to do data selection.
Some approaches use signals or features produced by LLMs, such as perplexity score~\cite{wang2023sass}, confidence levels~\cite{wang2022self,huang2023large}, and logits~\cite{pace2024west}, as criteria for constructing data selectors.
Others directly prompt the LLMs for this task.
For instance,~\citet{lu2023self} query the target LLM to assess the quality of generated samples.
\citet{kim2023soda} leverage ChatGPT to determine if the social commonsense knowledge is appropriately conveyed in the synthetic dialogues.
Additionally, there are also works that adopt the LLMs to rank multiple candidate annotations and utilize the top ones in the subsequent stages~\cite{jeronymo2023inpars,li2024dalk}.
In pairwise feedback synthesis,~\citet{tong2024optimizing} task the base LLM with judging whether one response genuinely surpasses another.
Besides,~\citet{jiang2024importance} demonstrate that filtering out correct but with high distribution shift extent (DSE) samples could also benefit the results of self-improvement.

% \end{document}

% \documentclass[../LLM4Annotate.tex]{subfiles}

% \begin{document}
\vspace{-1mm}
\section{LLM-Generated Annotations Utilization}
\vspace{-2mm}
LLM-generated annotations provide a valuable resource of labeled data for NLP models in different stages. Hereby we explore the methods for utilizing and learning with LLM-Generated Annotations.
\vspace{-4mm}
\subsection{Supervised Fine-Tuning}

Supervised fine-tuning can effectively enhance models' specific capabilities or knowledge.
In this section, we discuss the utilization of generated annotation for supervised fine-tuning.

\noindent\textbf{Self-Evolution.} \citet{huang2023large} first propose the concept of self-improve that utilizes LLMs as both data annotators and learnable models and iteratively fine-tune LLMs in their self-annotated data.
\citet{wang2023self} also tune a GPT3 in the instruction tuning dataset to improve its zero-shot generalization capability.
To foster LLMs' evolution,~\citet{lu2023self} iteratively fine-tune the LLMs in self-refined synthetic responses.
To mitigate the distribution gap between task datasets and the LLMs,~\citet{yang2024self} use self-distillation which guides fine-tuning with a distilled dataset generated by the model itself.
Both~\citet{chen2024self} and~\citet{cheng2024self} introduce a self-play mechanism, where the LLM refines its capability by playing against instances of itself.
Moreover,~\citet{wang2024self} demonstrate that the reasoning abilities of small-scale LMs can be enhanced through self-training, a process where models learn from their own outputs. 

\noindent\textbf{Distill Smaller Models.} For efficiency issues, many studies aim to use the data generated by a large and powerful LLM to train a flexible and affordable smaller model.
For a better instruction-following ability, many medium and small-sized LLMs are trained on the synthetic dataset produced by larger LLMs~\cite{taori2023stanford,chiang2023vicuna,xu2023wizardlm}.
In classification tasks,~\citet{meng2022generating,meng2023tuning,wang2023noise} augment the original datasets and train smaller bidirectional attention models on them.
To foster models' reasoning ability, many studies tune smaller models with synthetic rationales collected from LLMs~\cite{wang2022pinto,shridhar2023distilling,liu2023logicot,kang2024knowledge}.
Other task-specific capabilities distillation from LLMs include dialogue generation~\cite{xu2023baize}, information extraction~\cite{josifoski2023exploiting,jeronymo2023inpars} and code generation~\cite{chaudhary2023code,roziere2023code}.
Moreover, LLMs have been proven to follow a scaling law in terms of their knowledge capacity.
Therefore, there is also a growing interest in distilling vertical and domain-specific knowledge from LLMs, including medicine~\cite{zhang2023huatuogpt,xiong2023doctorglm}, finance~\cite{zhang2023xuanyuan} and science~\cite{luo2023wizardmath,zhao2024gimlet}, to smaller models.
\vspace{-3mm}
\subsection{Alignment Tuning}
\vspace{-1mm}

Alignment tuning methods, like RLHF~\cite{ouyang2022training}, aim to align the output of LLMs with human intentions, ensuring they are helpful, ethical, and reliable.
Synthetic data produced by LLMs are widely adopted in these alignment approaches for reward modeling and policy training.

\noindent\textbf{Reward Modeling.}
LLMs-generated annotations can be used to train or refine the reward model for better alignment.
\citet{xu2023contrastive} propose a data curriculum method that leverages the pairwise feedback from LLMs to calculate the sample difficulty level and smooth LLMs' learning from simple ones to hard ones.
\citet{kim2023aligning} design reward model guided self-play to iteratively improve the reward model with synthesized data generated by the policy model.
\citet{pace2024west} propose to maximize the probability of correctly labeling a pair of on-policy responses to a given query according to the base preference model.
In robotics,~\citet{zeng2024learning} learns a reward function from scratch using the LLMs' feedback.
With synthetic data pair,~\citet{Sun2023SALMONSW} train an instructable reward model to generate reward scores based on arbitrary human-defined principles.

\noindent\textbf{Policy Training.}
While many direct alignment methods~\cite{rafailov2024direct,zhao2023slic} have emerged recently, some works directly explore the use of annotated feedback for policy training.
One common strategy is to directly apply DPO with the synthetic pairwise feedback produced by LLMs~\cite{yuan2024self,zhang2024self,lee2024aligning,tong2024optimizing,lee2024reinforcement,guo2024direct}.
Besides,~\citet{gulcehre2023reinforced,dong2023raft} leverage a pre-trained reward model to filter low-quality synthetic data and iteratively tune LLMs with growing datasets.
\citet{wang2024step} propose a bootstrapping self-alignment method to repeatly utilize the synthetic data.
\citet{liu2024mixture} introduce the Mixture of insighTful Experts (MoTE) architecture, which applies the mixture of experts to enhance each component of the synthetic response, markedly increasing alignment efficiency.
With the reasoning pairwise feedback generated by LLM itself,~\citet{pang2024iterative} use a modified DPO loss with an additional negative log-likelihood term to tune the LLM.

\vspace{-0.05in}
\subsection{Inference}

\noindent\textbf{In-Context Learning.}
In-context Learning (ICL) consists of three components: a task description (or prompt), several in-context samples (or demonstration), and the test case that needs to be inferred.
Current studies have applied the annotations and data generated by LLMs in all these components for refining or augmenting.
\citet{zhou2022large} first showed that with a well-designed pipeline, LLMs can be human-level prompt engineers to generate accurate task descriptions.
Following them, ~\citet{yang2023auto,liempowering} conduct augmentation and expansion to the original task prompt, making it more detailed for LLMs to follow.
Demonstration augmentation~\cite{kim2022self,li2023human,chen2023self,he2024self} is another useful skill to enrich and diversify the provided demonstrations, especially when the labeled data is limited.
For the test sample, one augmentation method is to leverage LLMs to rephrase it once~\cite{deng2023rephrase} or multiple times~\cite{li2023dail,yang2024just}.
Other works study how to polish the original test sample~\cite{xi2023self} or decompose it into several sub-questions~\cite{wang2024self}.

\noindent\textbf{Reasoning.} Reasoning plays a crucial role in enhancing the quality and accuracy of the content generated by LLMs.
One efficient manner to boost LLMs' reasoning with self-generated annotation is to provide the generated rationale directly before outputting the final answer/ response~\cite{kojima2022large}.
To improve LLMs' performance with multiple reasoning pathways, majority voting~\cite{wang2022self,chen2023universal} and elimination~\cite{tong2023eliminating,balepur2023s,ma2023poe} are adopted to decide the final answer among several possible candidates.
Post-hoc editing and refining~\cite{madaan2024self,tong2024can} is another well-studied direction to utilize textual feedback and analysis for improving LLMs' reasoning capabilities.
Additionally, utilization of LLMs-generated annotations sometimes requires additional domain tools.
For example,~\citet{chen2023program} use a program interpreter in program-of-thought (PoT) to execute the generated program and convert it to a specific answer.
\citet{besta2024graph} design a prompter to Build a prompt to be sent to the LLM and a parser to extract information from LLM thought.
In tree-of-thought (ToT),~\citet{hao2023reasoning,yao2024tree} build an additional state evaluator by designing specific prompts and repurposing the base LLM.

% \end{document}

% \documentclass[../LLM4Annotate.tex]{subfiles}

% \begin{document}
\vspace{-0.1in}
\section{Societal Impact and Future Work}
\vspace{-2mm}
In this section, we outline LLM annotation challenges, including societal implications, technical concerns, and bias propagation. 
% Addressing these is vital for advancing LLM annotations.

\vspace{-2mm}
\subsection{Ethics Consideration}
One critical concern of LLM-generated annotations is the ethics consideration, especially in high-stakes decision-making tasks like finance~\cite{yang2023fingpt}, jurisprudence~\cite{cui2023chatlaw}, and healthcare~\cite{eloundou2023gpts}. Despite the efficiency of LLM annotation, the lack of human insight may lead to biased and unfair results~\cite{wu2023bloomberggpt, abid2021persistent, cheng2021socially, li2023survey, beigi2024model, das2024investigating, shimabucoro2024llm}. Moreover, LLMs make human annotator roles redundant, potentially increasing social disparities~\cite{dillion2023can}. Future studies should harmonize technological advancements with societal consequences, including considering social implications, ensuring ethical use, promoting fairness, and maintaining transparency. 
%For example, continuous monitoring and updates to LLMs, along with stakeholder engagement, are necessary. These measures will improve model fairness and reliability, ensuring that LLM applications for data annotation serve the public interest and adhere to ethical standards.
\vspace{-2mm}
\subsection{Challenges and Future Work}

\textbf{Model Collapse.}
Model collapse refers to the gradual performance decrease of an LLM trained on the outputs of other LLMs~\cite{sun2023chatgpt, Gunasekar2023TextbooksAA, hsieh2023distilling, honovich2022unnatural, vicuna2023, koala_blogpost_2023,huang2024authorship}. It is unavoidable since LLM-generated data is occupying the information ecosystem. The imitation model often replicates stylistic elements without achieving the factual precision of superior models~\cite{Gudibande2023TheFP, Shumailov2023TheCO}. This divergence is caused by \textit{statistical approximation error} from limited sample sizes and \textit{functional approximation error} from constrained model capacity. Both errors tend to amplify through successive training cycles~\cite{Alemohammad2023SelfConsumingGM}. 
% Using LLM-generated annotations with these inaccuracies can lead to data contamination, undermining LLMs' trustworthiness and impacting their utility in critical applications.

\noindent\textbf{Potential Solution.} It is important to ensure that the training data is diverse and high-quality, with a significant proportion of human-generated content. \citet{gerstgrasser2024model} avoid model collapse by accumulating real and machine-generated data. This method maintains data diversity, preventing performance degradation across different LLMs.

\noindent\textbf{Hallucinations.}
Hallucinations in LLMs significantly undermine the integrity and reliability of their generated annotations~\cite{alkaissi2023artificial, azamfirei2023large,chaudhary2024brainstorm}. Hullicinated outputs detached from factual information can cause the proliferation of misinformation~\cite{jiang2024disinformation, chen2023can,chencan,huang2024can}. Addressing hallucinations requires refining the training process and implementing validation mechanisms for annotations through automated and manual verification~\cite{liao2023ai, pan2023risk, bian2023drop}. Moreover, the inherent opacity of LLMs complicates efforts to investigate the causes of hallucinations.

\noindent\textbf{Potential Solution.} \citet{yang2023new} addresses hallucinations in LLMs with the Reverse Validation method, detecting hallucinations at the passage level by constructing a query from the response and checking for a match within the LLM's internal knowledge. \citet{bertaglia2023closing} uses Chain-of-Thought (CoT) prompting and explanation generation, where CoT prompting produces explanations for predictions, ensuring logical and verifiable outputs. \citet{li2023coannotating} proposes the CoAnnotating framework, which uses uncertainty-guided work allocation between humans and LLMs, applying self-evaluation and entropy metrics to assess reliability and distribute tasks effectively. \citet{zendel2024enhancing} propose a human-LLM connotation process for better annotation quality.

% \subsubsection{Efficiency of LLMs}
% Efficiency in LLMs is crucial because of their growing size and complexity, which demand substantial computational resources and memory. Efficient models reduce inference latency, which is vital for real-time applications. They also lower energy consumption, supporting sustainable AI practices. Additionally, they cut down operational costs in cloud environments, making AI more cost-effective for businesses and researchers. Efficiency techniques for LLMs, such as pruning, compression, and distillation, are critical for deploying these models in resource-constrained environments. \cite{xu2023knowledge} introduces CLINGEN, a resource-efficient method for generating synthetic clinical text using LLMs. By employing a knowledge infusion technique, it integrates clinical knowledge to enhance data generation and reduce the need for large labeled datasets. \cite{yu2024large} uses attribute-based prompting (AttrPrompt) for efficiency in LLMs. AttrPrompt generates diverse training data with varied attributes like length, style, and location, enhancing data quality and reducing biases. \cite{peng2023generating} introduces Chain-of-Thought Attribute Manipulation (CoTAM) for efficient data augmentation in LLMs. CoTAM enhances efficiency by decomposing text into attributes, proposing changes to specific attributes, and reconstructing sentences with these modifications.

\noindent\textbf{Efficiency of LLMs.}
Efficiency in LLMs is crucial due to their growing size and complexity, which demand substantial computational resources~\cite{wong2024efficiency}. Efficient models reduce inference latency, vital for real-time applications, lower energy consumption for sustainable AI practices, and cut operational costs in cloud environments, making AI more cost-effective for researchers. Efficiency techniques for LLMs, such as pruning, compression, and distillation, are critical for deploying these models in resource-constrained environments.

\noindent\textbf{Potential Solution.} Pruning is an efficient technique to reduce the number of parameters in an LLM. For example,~\citet{ma2023llm} selectively removes redundant neurons based on gradient information while preserving most of the LLM's capability. Mixture of Experts (MoE) is another promising technique that leverages a set of expert sub-models, where only a subset of these experts is activated for any given input~\cite{artetxe2021efficient}. Researchers also adopt LLM Quantization to reduce the precision of the numbers used to represent a model's parameters~\cite{xiao2023smoothquant}. Instead of using 32-bit floating-point numbers, a quantized model might use 16-bit floats, 8-bit integers, or even lower precision. These techniques can be combined with each other to achieve further efficiencies.
% Instead of fine-tuning all weights of the LLM, parameter-efficient fine-tuning techniques, such as LoRA~\cite{hu2021lora} and prompt tuning~\cite{lester2021power}, only optimize a small portion of the model parameters while keeping the rest frozen to perform specific downstream tasks.

% \noindent\textbf{Potential Solution.} \cite{xu2023knowledge} introduces CLINGEN, a resource-efficient method for generating synthetic clinical text using LLMs. By employing a knowledge infusion technique, it integrates clinical knowledge to enhance data generation. \cite{yu2024large} proposes AttrPrompt, which generates diverse training data with different attributes like length, style, and location, enhancing data quality and reducing biases. \cite{peng2023generating} uses Chain-of-Thought Attribute Manipulation to decompose text into attributes and reconstruct sentences.

% \noindent\textbf{Compounding Error in Model Imitation.}

% \textbf{Harmful Annotations.}
% The proficiency of LLMs in inferring annotations isn't devoid of risks. Malicious actors can exploit well-crafted prompts, leading to LLM safety issues~\cite{wang2023robustness, zhu2023promptbench}. Such misuse encompasses membership inference attacks and copyright violations~\cite{carlini2021extracting, carlini2022quantifying}.

% \end{document}

% \documentclass[../LLM4Annotate.tex]{subfiles}

% \begin{document}
\vspace{-2mm}
\section{Conclusion}
The exploration of LLMs for data annotation and synthesis has revealed an exciting frontier in NLP, presenting novel solutions to longstanding challenges like data scarcity, and enhancing annotation quality and process efficiency. This survey meticulously reviews methodologies, applications, and hurdles associated with LLM employment, including detailed taxonomy from annotation generation to utilization. It evaluates the effects of LLM-generated annotations on training machine learning models while addressing both technical and ethical concerns like bias and societal ramifications. Highlighting our novel taxonomy of LLM methodologies, strategies for utilizing LLM-generated annotations, and a critical discussion on the challenges, this work aims to steer future progress in this crucial area. Additionally, we introduce a comprehensive categorization of techniques and compile extensive benchmark datasets to support ongoing research endeavors, concluding with an examination of persistent challenges and open questions, paving the way for future investigative pursuits in the domain.

% \end{document}

% \documentclass[../LLM4Annotate.tex]{subfiles}

% \begin{document}
% 1. Focus more on discriminative tasks
% 2. 
\section*{Limitations}
\noindent\textbf{Sampling Bias and Hallucination.} LLMs can display sampling bias, leading to incorrect or ``hallucinated'' data, impacting the reliability and quality of annotations for discriminative tasks.

\noindent\textbf{Social Bias and Ethical Dilemmas.} The inherent biases in training data can be perpetuated and amplified by LLMs, leading to ethical concerns and the propagation of social biases through annotated data. This is particularly problematic in tasks requiring fairness and impartiality.

\noindent\textbf{Dependence on High-Quality Data.} LLMs' usefulness in generating annotations depends on large, high-quality datasets. But curating these datasets is labor-intensive, posing a scalability challenge for LLM-based annotation efforts.

\noindent\textbf{Complexity in Tuning and Prompt Engineering.} Successfully leveraging LLMs for data annotation requires sophisticated prompt engineering and fine-tuning techniques. This can pose a barrier to entry for practitioners and researchers without extensive expertise in NLP and machine learning.

\noindent\textbf{Generalization and Overfitting} While LLMs can be powerful tools for annotation, there's a risk of overfitting to the training data, limiting their ability to generalize to unseen data or different contexts. This is a critical limitation for discriminative tasks where the goal is to develop models that perform well across diverse datasets and domains.

\noindent\textbf{Computational and Resource Requirements.} The training and deployment of state-of-the-art LLMs for data annotation require substantial computational resources, which may not be accessible to all researchers and organizations, thereby limiting widespread adoption.

\section*{Acknowledgements} The material in this presentation is supported by the National Science Foundation (NSF) under grants IIS-2229461, and the U.S. Department of Homeland Security under Grant Award Number, 17STQAC00001-08-00 and the U.S. Office of Naval Research (ONR) under grant N00014-21-1-4002. Lu Cheng is supported by the National Science Foundation (NSF) Grant \#2312862, NIH \#R01AG091762, and a Cisco gift grant. 
The views and conclusions contained in this document are those of the authors and should not be interpreted as necessarily representing the official policies, either expressed or implied, of the U.S. Department of Homeland Security and the National Science Foundation.

\bibliography{custom}
\bibliographystyle{acl_natbib}

\appendix

\section{LLM-assisted Tools and Software for Annotation}
\label{app:tool}

LLM-assisted annotation tools and software are invaluable resources designed specifically to facilitate the annotation process for various NLP tasks.
One of their primary attributes is an intuitive and user-friendly interface, allowing engineers and even non-technical annotators to easily work with complex textual data. These tools are built to support numerous annotation types, from simple binary labels to more intricate hierarchical structures.
The main goal of these tools is to simplify the labeling process, enhance the quality of the labels, and boost overall productivity in data annotation.

    \begin{figure*}[ht]
        \centering
        \includegraphics[width=0.8\linewidth]{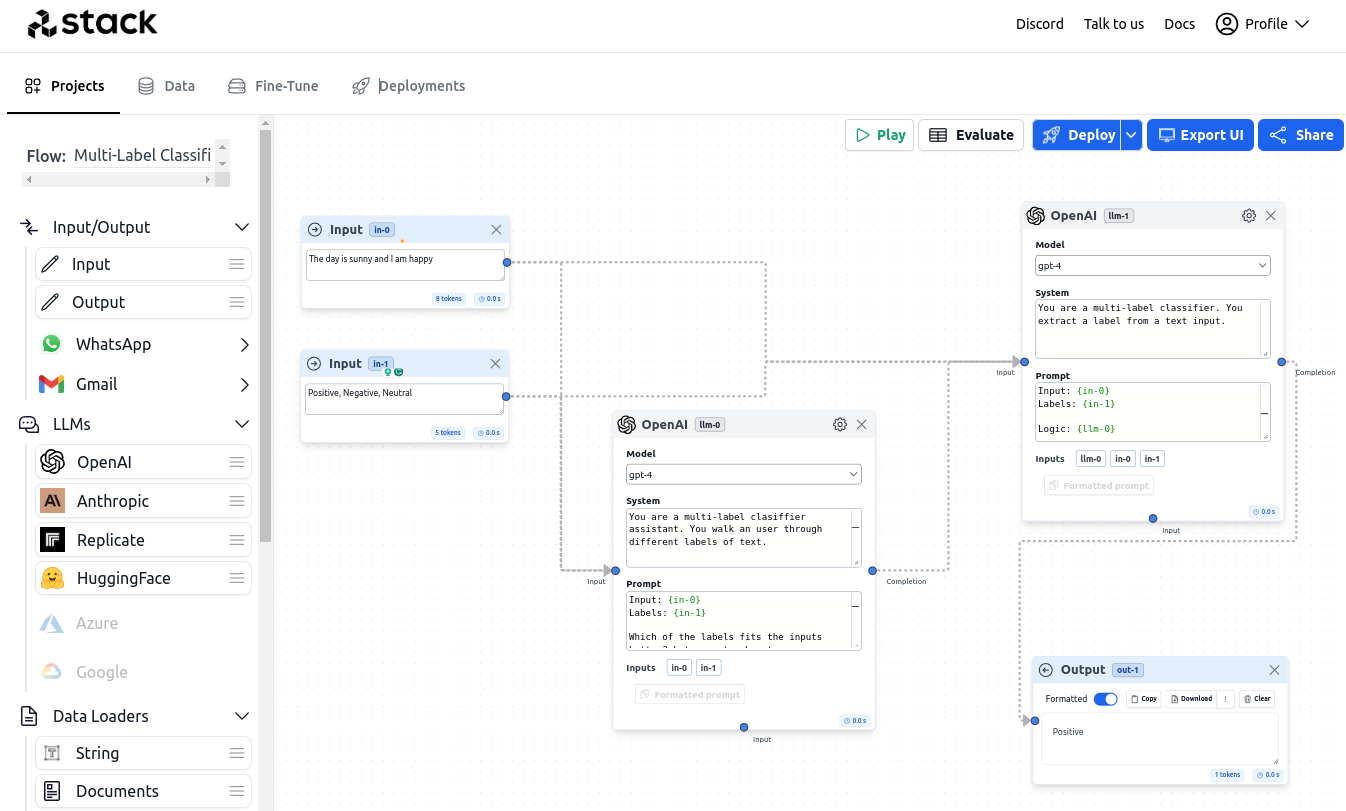}
        \caption{Stack AI dashboard. They provide a visual interface for users to design and track the AI workflow.}
        \label{fig::stackai} 
    \end{figure*}

Below, we will present a selection of the libraries and tools that support Large Language Models for the annotation process:
\begin{itemize}

    \item \textbf{LangChain}: LangChain~\cite{langchain} is an open-source library\footnote{As of now, available only in JavaScript/TypeScript and Python languages.} that offers an array of tools designed to facilitate the construction of LLM-related pipelines and workflows. This library specifically provides large language models with agents in order to interact effectively with their environment as well as various external data sources. Therefore, providing dynamic and contextually appropriate responses that go beyond a single LLM call. 
    
    In terms of the annotation process, their power mostly lies in the facilitation of annotation through the creation of a modularized structure called \textit{chain}.
    In the chaining technique, a complex problem is broken down into smaller sub-tasks. The results obtained from one or more steps are then aggregated and utilized as input prompts for subsequent actions in the chain.
    
    \item \textbf{Stack AI}: Stack AI~\cite{stackai} is a paid service that offers an AI-powered data platform. It is designed explicitly for automating business processes allowing them to maximize efficiency. The essence of their platform lies in their ability to \textit{visually} design, test, and deploy AI workflows through smooth integration of Large Language Models. Their user-friendly graphical interface (Figure~\ref{fig::stackai}) allows the users to create apps and workflows related to diverse tasks from content creation and data labeling to conversational AI apps and document processing. Moreover, Stack AI utilizes weakly supervised machine learning models to expedite the data preparation process.

    \begin{figure}[ht]
        \centering
        \includegraphics[width=\linewidth]{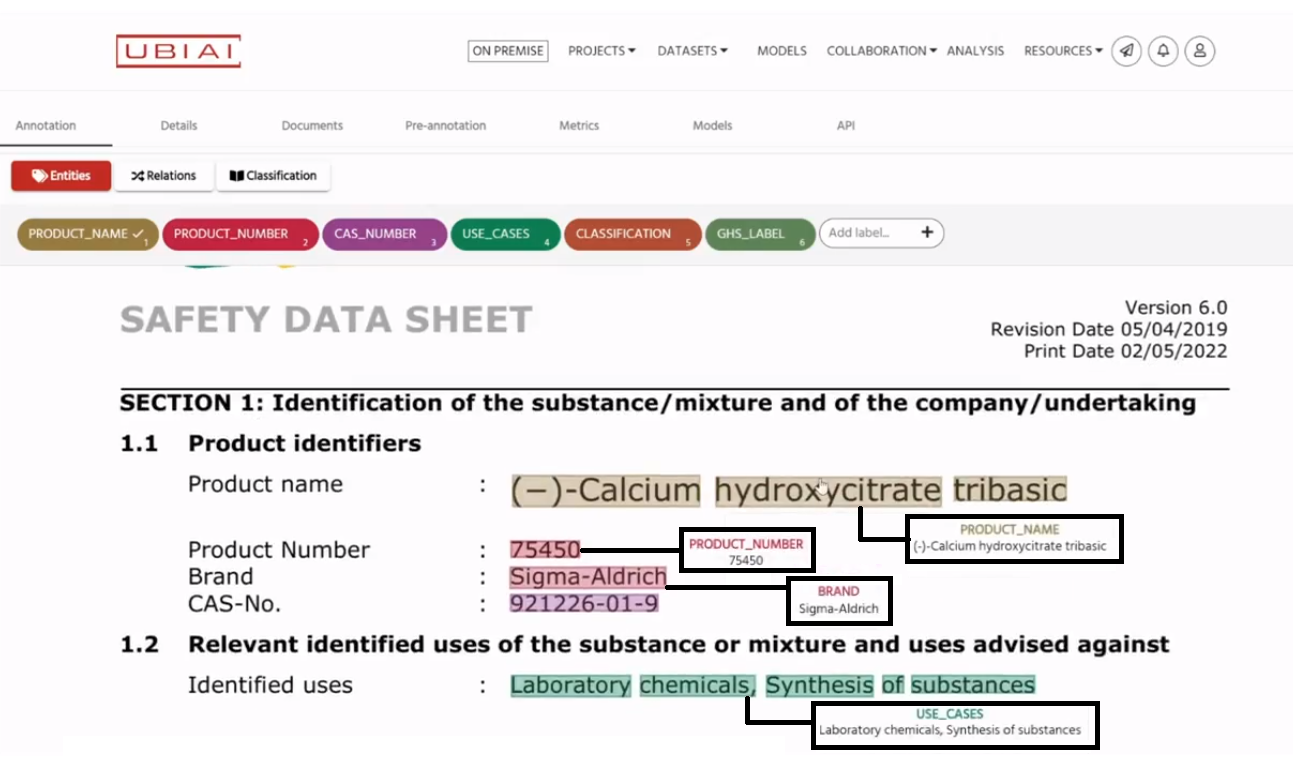}
        \caption{UBIAI annotation result on a pdf document. All the entities in the text of the document have been identified, annotated, and color-coded based on the type. This image has been borrowed from the videos provided in the UBIAI documentation~\cite{ubiai}.}
        \label{fig::UBIAI} 
    \end{figure}
        
    \item \textbf{UBIAI}:
    UBIAI~\cite{ubiai} is a paid annotation tool that offers multilingual cloud-based solutions and services in Natural Language Processing. The company aims to aid users in extracting valuable insights from unstructured documents. This tool not only provides a user interface that facilitates manual labeling but also offers several auto-labeling functionalities such as LLM-assisted zero- and few-shot labeling and model-assisted labeling. They also provide integration to various models on huggingface~\cite{wolf2020huggingfaces} as well as an environment to fine-tune different models on the user's labeled data.

    \item \textbf{Prodigy}: Prodigy~\cite{Prodigy2018}, designed by the creators of spaCy library~\cite{spacy2}, offers rule-based, statistical models, and LLM-assisted methods for annotation. This tool provides easy, flexible, and powerful annotation options such as named entity recognition, span categorization, and classification/labeling for different modalities including text, audio, and vision.
    Moreover, it can be easily integrated with large language models which are capable of zero- or few-shot learning, while also offering services and quantifiable methods for crafting prompts to address any noisy outcomes. This tool is not open-source.

\end{itemize}

\section{Acknowledgment of AI Assistance in Writing and Revision}
We utilized ChatGPT-4 for revising and enhancing sections of this paper.

%%%%%%%%%%%%%%%%%%%%%%%%%%%%%%%%%%%%%%%%%%%%%%%%%%%%%%%%%%%%%%%
\section{Collections of Papers on LLM for Data Annotation}
\label{app:table}
This collection of tables provides a concise overview of using Large Language Models (LLMs) for data annotation, including state-of-the-art techniques, methodologies, and practical applications. Table~\ref{tab:LLM-Based Annotation Generation-1} and Table~\ref{tab:LLM-Based Annotation Generation-2} lists significant papers on LLM-based data annotation, detailing their methods, core technologies, publication venues, and links to resources. Table~\ref{tab:LLM-Generated Annotation Assessment} focuses on assessment and filtering of LLM-generated annotations. Tables~\ref{tab:LLM-Generated Annotation Utilization} explore strategies for learning with LLM-generated annotations, covering supervised fine-tuning, alignment tuning and inference. Each table clearly outlines the data type, backbone, computational cost, venues, and available resources, serving as a guide to the latest in LLM-driven data annotation and its implications for the future of automated data processing and machine learning research.

\begin{table*}[t]
\centering
% \scriptsize
\resizebox{1\textwidth}{!}{
\begin{tabular} {c|c|c|c|c|c}
\rowcolor{Gray}
\toprule
Paper & Data Type & Backbone & Annotation Cost & Venue & Code/Data Link \\
\midrule

\rowcolor{Gray}
\multicolumn{6}{c}{Instruction \& Response}\\
\midrule

\shortstack{GPT3Mix: Leveraging Large-scale Language Models for Text Augmentation$^{[1]}$} & Instruction & GPT-3 & \begin{tabular}[c]{@{}c@{}}API Calling,\\ 300 tokens per sample\end{tabular} & EMNLP'21 & \href{https://github.com/naver-ai/hypermix}{Link}\\
\midrule

\shortstack{SELF-INSTRUCT: Aligning Language Models with Self-Generated Instructions$^{[2]}$} & Instruction \& Response & GPT-3 & \begin{tabular}[c]{@{}c@{}} API Calling,\\ \$600 for entire dataset \end{tabular} & ACL'23 & \href{https://github.com/yizhongw/self-instruct}{Link}\\
\midrule

\shortstack{Tuning Language Models as Training Data Generators for Augmentation-Enhanced Few-Shot Learning$^{[3]}$} & Instruction &  CTRL & \begin{tabular}[c]{@{}c@{}}  Model Training,\\ Nvidia A100 GPUs,\\ 10 minutes per task   \end{tabular}  & ICML'23 & \href{https://github.com/yumeng5/FewGen}{Link}\\
\midrule

\shortstack{SASS: SELF-ALIGNMENT WITH SEMI-SUPERVISED INSTRUCTION DATA GENERATION$^{[4]}$} & Instruction &  LLaMA  & \begin{tabular}[c]{@{}c@{}}  Model Training,\\ Nvidia A100 GPUs    \end{tabular}& OpenRview'24 & Not Available \\
\midrule

\shortstack{DAIL: Data Augmentation for In-Context Learning via Self-Paraphrase$^{[5]}$} & Instruction & ChatGPT & API Calling & Arxiv'23 & Not Available \\
\midrule

\shortstack{LongForm: Effective Instruction Tuning with Reverse Instructions$^{[6]}$} & Instruction & GPT-3 & PI Calling & ICLR'24 & \href{https://github.com/akoksal/LongForm}{Link}\\
\midrule

\shortstack{Large Language Model as Attributed Training Data Generator: A Tale of Diversity and Bias$^{[7]}$} & Instruction & ChatGPT & API Calling & NeurIPS'23 & \href{https://github.com/yueyu1030/AttrPrompt}{Link}\\
\midrule

\shortstack{SELF-QA: Unsupervised Knowledge Guided Language Model Alignment$^{[8]}$} & Instruction \& Response &  BLOOM & Model Inference & Arxiv'23 & Not Available \\
\midrule

\shortstack{LARGE LANGUAGE MODELS CAN SELF-IMPROVE$^{[9]}$} & Response &  PaLM-540B & Model Inference & EMNLP'23 & Not Available \\
\midrule

\shortstack{Self-Distillation Bridges Distribution Gap in Language Model Fine-Tuning$^{[10]}$} & Response & LLaMA-2 & Model Inference & ACL'24 & \href{https://github.com/sail-sg/sdft}{Link}\\
\midrule

\shortstack{Mixture of insighTful Experts (MoTE): The Synergy of Thought Chains and Expert Mixtures in Self-Alignment$^{[11]}$} & Response & Alpaca & Model Inference & Arxiv'24 & Not Available\\
\midrule

\shortstack{Human-Instruction-Free LLM Self-Alignment with Limited Samples$^{[12]}$} & Instruction \& Response & Multiple LLMs & \begin{tabular}[c]{@{}c@{}}  Model Inference,\\ single NVIDIA A100 80G GPU   \end{tabular} & Arxiv'24 & Not Available\\
\midrule

\shortstack{Principle-Driven Self-Alignment of Language Models from Scratch with Minimal Human Supervision$^{[13]}$} & Response & LLaMA & Model Inference & NeurIPS'23 & \href{https://github.com/IBM/Dromedary}{Link}\\
\midrule

\shortstack{Step-On-Feet Tuning: Scaling Self-Alignment of LLMs via Bootstrapping$^{[14]}$} & Response & LLaMA-2 & Model Inference & Arxiv'24 & Not Available\\
\midrule

\shortstack{Assessing Empathy in Large Language Models with Real-World Physician-Patient Interactions$^{[15]}$} & Response & LLaMA & Model Inference & Arxiv'24 & Not Available\\
\midrule

\rowcolor{Gray}
\multicolumn{6}{c}{Rationale}\\
\midrule

\shortstack{Large Language Models are Zero-Shot Reasoners$^{[16]}$} & Rationale - CoT & Multiple LLMs & API Calling & NeurIPS'22 & Not Available \\
\midrule

\shortstack{Tree of Thoughts: Deliberate Problem Solving with Large Language Models$^{[17]}$} & Rationale - Tree & GPT-4 & API Calling, \$0.74 per sample & NeurIPS'22 & \href{https://github.com/princeton-nlp/tree-of-thought-llm}{Link}\\
\midrule

\shortstack{Reasoning with Language Model is Planning with World Model$^{[18]}$} & Rationale - Tree &  LLaMA & \begin{tabular}[c]{@{}c@{}}  Model Inference,\\ 4×24 GB NVIDIA A5000 GPUs   \end{tabular} & EMNLP'23 & \href{https://github.com/maitrix-org/llm-reasoners}{Link}\\
\midrule

\shortstack{Graph of Thoughts: Solving Elaborate Problems with Large Language Models$^{[19]}$} & Rationale - Graph & GPT-3.5 & API Calling & AAAI'24 & \href{https://github.com/spcl/graph-of-thoughts}{Link}\\
\midrule

\shortstack{Beyond Chain-of-Thought, Effective Graph-of-Thought Reasoning in Language Models$^{[20]}$} & Rationale - Graph & GPT-3 & API Calling & Arxiv'23 & \href{https://github.com/Zoeyyao27/Graph-of-Thought}{Link}\\
\midrule

\shortstack{CHAIN-OF-TABLE: EVOLVING TABLES IN THE REASONING CHAIN FOR TABLE UNDERSTANDING$^{[21]}$} & Rationale - Table & Multiple LLMs & API Calling \& Model Inference & ICLR'24 & Not Available\\
\midrule

\shortstack{Program of Thoughts Prompting: Disentangling Computation from Reasoning for Numerical Reasoning Tasks
$^{[22]}$} & Rationale - Program & Multiple LLMs & API Calling \& Model Inference & TMLR'23 & Not Available\\
\midrule

\shortstack{The Art of SOCRATIC QUESTIONING: Recursive Thinking with Large Language Models$^{[23]}$} & Rationale - Reversion & ChatGPT & \begin{tabular}[c]{@{}c@{}}  API Calling,\\ 9.22 calls per sample   \end{tabular} & EMNLP'23 & \href{https://github.com/VT-NLP/SOCRATIC-QUESTIONING}{Link}\\
\midrule

\shortstack{Interpreting Pretrained Language Models via Concept Bottlenecks$^{[24]}$} & Rationale - Concept & ChatGPT & API Calling & PAKDD'24 & \href{https://github.com/Zhen-Tan-dmml/CBM_NLP}{Link}\\
\midrule

\shortstack{PINTO: FAITHFUL LANGUAGE REASONING USING PROMPT-GENERATED RATIONALES$^{[25]}$} & Rationale - CoT & GPT-neox & Model Inference & ICLR'23 & \href{https://github.com/wangpf3/pinto-faithful-language-reasoning}{Link}\\
\midrule

\shortstack{SCOTT: Self-Consistent Chain-of-Thought Distillation$^{[26]}$} & Rationale - CoT &  GPT-neox & Model Inference & ACL'23 & \href{https://github.com/wangpf3/consistent-CoT-distillation}{Link}\\
\midrule

\shortstack{LogiCoT: Logical Chain-of-Thought Instruction Tuning$^{[27]}$} & Rationale - CoT & GPT-4 & API Calling & EMNLP'23 & Not Available \\
\midrule

\shortstack{Distilling Reasoning Capabilities into Smaller Language Models$^{[28]}$} & Rationale - CoT & GPT-3 & API Calling & ACL'23 & Not Available\\
\midrule

\shortstack{Knowledge-Augmented Reasoning Distillation for Small Language Models in Knowledge-Intensive Tasks
$^{[29]}$} & Rationale - CoT & ChatGPT & API Calling & NeurIPS'23 & \href{https://github.com/Nardien/KARD}{Link}\\
\midrule

\shortstack{Making Pre-trained Language Models Better Few-shot Learners$^{[30]}$} & Rationale - Diverse Thinking & GPT-3 & API Calling & ACL'21 & \href{https://github.com/princeton-nlp/LM-BFF}{Link}\\
\midrule

\shortstack{SELF-CONSISTENCY IMPROVES CHAIN OF THOUGHT REASONING IN LANGUAGE MODELS$^{[31]}$} & Rationale - Diverse Thinking & Multiple LLMs & API Calling \& Model Inference & ICLR'23 & Not Available\\
\midrule

\shortstack{UNIVERSAL SELF-CONSISTENCY FOR LARGE LANGUAGE MODEL GENERATION$^{[32]}$} & Rationale - Diverse Thinking & Multiple LLMs & API Calling & Arxiv'23 & Not Available \\
\midrule

\shortstack{Plan, Verify and Switch: Integrated Reasoning with Diverse X-of-Thoughts$^{[33]}$} & Rationale - Diverse Thinking & ChatGPT & API Calling & EMNLP'23 & \href{https://github.com/tengxiaoliu/XoT}{Link}\\
\midrule

\shortstack{Eliminating Reasoning via Inferring with Planning: A New Framework to Guide LLMs’ Non-linear Thinking$^{[34]}$} & Rationale - Elimination & PaLM2 & API Calling & Arxiv'23 & Not Available\\
\midrule

\shortstack{It’s Not Easy Being Wrong: Large Language Models Struggle with Process of Elimination Reasoning$^{[35]}$} & Rationale - Elimination & Multiple LLMs & API Calling & ACL'24 & \href{https://github.com/nbalepur/PoE}{Link}\\
\midrule

\shortstack{POE: Process of Elimination for Multiple Choice Reasoning$^{[36]}$} & Rationale - Elimination & FLAN-T5 & Model Inference & EMNLP'23 & \href{https://github.com/KasMasVan/PoE}{Link}\\
\midrule

\shortstack{Exchange-of-Thought: Enhancing Large Language Model Capabilities through Cross-Model Communication
$^{[37]}$} & Rationale - Collaboration & ChatGPT & API Calling & EMNLP'23 & Not Available\\
\midrule

\shortstack{Encouraging Divergent Thinking in Large Language Models through Multi-Agent Debate$^{[38]}$} & Rationale - Collaboration & ChatGPT & API Calling & Arxiv'23 & \href{https://github.com/Skytliang/Multi-Agents-Debate}{Link}\\
\midrule

\shortstack{Towards Reasoning in Large Language Models via Multi-Agent Peer Review Collaboration$^{[39]}$} & Rationale - Collaboration & ChatGPT & API Calling & Arxiv'23 & \href{https://github.com/HITsz-TMG/Multi-agent-peer-review}{Link}\\
\midrule

\shortstack{DYNAMIC LLM-AGENT NETWORK: AN LLM-AGENT COLLABORATION FRAMEWORK WITH AGENT TEAM OPTIMIZATION$^{[40]}$} & Rationale - Collaboration & ChatGPT & \begin{tabular}[c]{@{}c@{}}     
 API Calling,\\ 16.5 calls per sample \end{tabular} & Arxiv'23 & \href{https://github.com/SALT-NLP/DyLAN}{Link}\\
\midrule

\rowcolor{Gray}
\multicolumn{6}{c}{Pair-wise Feedback}\\
\midrule

\shortstack{Constitutional AI: Harmlessness from AI Feedback$^{[41]}$} & Pairwise Feedback & Multiple LLMs & Model Inference & Arxiv'22 & \href{https://github.com/anthropics/ConstitutionalHarmlessnessPaper}{Link}\\
\midrule

\shortstack{RLAIF: Scaling Reinforcement Learning from Human Feedback with AI Feedback$^{[42]}$} & Pairwise Feedback & PaLM-2 & \begin{tabular}[c]{@{}c@{}} Model Inference,\\ \$0.67 per sample     \end{tabular}& Arxiv'23 & Not Available\\
\midrule

\shortstack{Self-Rewarding Language Models$^{[43]}$} & Pairwise Feedback & LLaMA-2 & Model Inference & Arxiv'24 &  Not Available \\
\midrule

\shortstack{SALMON: SELF-ALIGNMENT WITH INSTRUCTABLE REWARD MODELS$^{[44]}$} & Pairwise Feedback & LLaMA-2 & Model Inference & ICLR'24 & \href{https://github.com/IBM/SALMON}{Link}\\
\midrule

\shortstack{Self-Alignment for Factuality: Mitigating Hallucinations in LLMs via Self-Evaluation$^{[45]}$} & Pairwise Feedback & LLaMA & Model Inference & Arxiv'24 & \href{https://github.com/zhangxy-2019/Self-Alignment-for-Factuality}{Link}\\
\midrule

\shortstack{West-of-N: Synthetic Preference Generation for Improved Reward Modeling$^{[46]}$} & Pairwise Feedback &  T5-XXL & Model Inference  & Arxiv'24 & Not Available\\
\midrule

\shortstack{Learning Reward for Robot Skills Using Large Language Models via Self-Alignment$^{[47]}$} & Pairwise Feedback & ChatGPT & API Calling & ICML'24 & \href{https://sites.google.com/view/rewardselfalign}{Link}\\
\midrule

\shortstack{Aligning Large Language Models through Synthetic Feedback$^{[48]}$} & Pairwise Feedback & LLaMA & Model Inference & EMNLP'23 & \href{https://github.com/naver-ai/almost}{Link}\\
\midrule

\shortstack{Optimizing Language Model’s Reasoning Abilities with Weak Supervision$^{[49]}$} & Pairwise Feedback &  LLaMA & Model Inference & Arxiv'24 & Not Available\\
\midrule

\shortstack{RLCD: REINFORCEMENT LEARNING FROM CONTRASTIVE DISTILLATION FOR LM ALIGNMENT$^{[50]}$} & Pairwise Feedback & LLaMA & Model Inference & ICLR'24 & \href{https://github.com/facebookresearch/rlcd}{Link}\\
\midrule

\shortstack{Automatic Pair Construction for Contrastive Post-training$^{[51]}$} & Pairwise Feedback & LLaMA & \begin{tabular}[c]{@{}c@{}}  Model Inference,\\ 16 Nvidia V100 GPUs   \end{tabular}& NAACL'24 & Not Available\\
\midrule

\shortstack{Reinforcement Learning from Reflective Feedback (RLRF): Aligning and Improving LLMs via Fine-Grained Self-Reflection$^{[52]}$} & Pairwise Feedback & LLaMA-2 & \begin{tabular}[c]{@{}c@{}}  Model Inference,\\ 16 Nvidia V100 GPUs   \end{tabular} & Arxiv'24 & Not Available\\
\midrule

\shortstack{Improving Language Model Reasoning with Self-motivated Learning$^{[53]}$} & Pairwise Feedback & LLaMA-2 & Model Inference & LREC'24 & Not Available\\
\midrule

\toprule
\end{tabular}
}
\begin{tablenotes}
\item 
\small{Note: $^{[1]}$\cite{yoo2021gpt3mix};
$^{[2]}$\cite{wang2023self};
$^{[3]}$\cite{meng2023tuning};
$^{[4]}$\cite{wang2023sass};
$^{[5]}$\cite{li2023dail};
$^{[6]}$\cite{koksal2024longform};
$^{[7]}$\cite{yu2024large};
$^{[8]}$\cite{zhang2023self};
$^{[9]}$\cite{huang2023large};
$^{[10]}$\cite{yang2024self};
$^{[11]}$\cite{liu2024mixture};
$^{[12]}$\cite{guo2024human};
$^{[13]}$\cite{sun2024principle};
$^{[14]}$\cite{wang2024step};
$^{[15]}$\cite{luo2024assessing};
$^{[16]}$\cite{kojima2022large};
$^{[17]}$\cite{yao2024tree};
$^{[18]}$\cite{hao2023reasoning};
$^{[19]}$\cite{besta2024graph};
$^{[20]}$\cite{yao2023beyond};
$^{[21]}$\cite{wang2024chain};
$^{[22]}$\cite{chen2023program};
$^{[23]}$\cite{qi2023art};
$^{[24]}$\cite{tan2023interpreting};
$^{[25]}$\cite{wang2022pinto};
$^{[26]}$\cite{wang2023scott};
$^{[27]}$\cite{liu2023logicot};
$^{[28]}$\cite{shridhar2023distilling};
$^{[29]}$\cite{kang2024knowledge};
$^{[30]}$\cite{gao2021making};
$^{[31]}$\cite{wang2022self};
$^{[32]}$\cite{chen2023universal};
$^{[33]}$\cite{liu2023plan};
$^{[34]}$\cite{tong2023eliminating};
$^{[35]}$\cite{balepur2023s};
$^{[36]}$\cite{ma2023poe};
$^{[37]}$\cite{yin2023exchange};
$^{[38]}$\cite{liang2023encouraging};
$^{[39]}$\cite{xu2023towards};
$^{[401]}$\cite{liu2023dynamic};
$^{[41]}$\cite{bai2022constitutional};
$^{[42]}$\cite{lee2023rlaif};
$^{[43]}$\cite{yuan2024self};
$^{[44]}$\cite{Sun2023SALMONSW};
$^{[45]}$\cite{zhang2024self};
$^{[46]}$\cite{pace2024west};
$^{[47]}$\cite{zeng2024learning};
$^{[48]}$\cite{kim2023aligning};
$^{[49]}$\cite{tong2024optimizing};
$^{[50]}$\cite{yang2023rlcd};
$^{[51]}$\cite{xu2023contrastive};
$^{[52]}$\cite{lee2024reinforcement};
$^{[53]}$\cite{feng2024improving}.}
\end{tablenotes}
\caption{A list of representative LLM-Based Annotation Generation (Instruction \& Response, Rationale, Pairwise Feedback) papers with open-source code/data.} 
\label{tab:LLM-Based Annotation Generation-1}
\end{table*}
%%%%%%%%%%%%%%%%%%%%%%%%%%%%%%%%%%%%%%%%%%%%%%%%%%%%%%%%%%%%%%%

\begin{table*}[t]
\centering
% \scriptsize
\resizebox{1\textwidth}{!}{
\begin{tabular} {c|c|c|c|c|c}
\rowcolor{Gray}
\toprule
Paper & Data Type & Backbone & Annotation Cost & Venue & Code/Data Link \\
\midrule

\rowcolor{Gray}
\multicolumn{6}{c}{Textual Feedback}\\
\midrule

\shortstack{SELF-REFINE: Iterative Refinement with Self-Feedback$^{[1]}$} & Textual Feedback & Multiple LLMs & API Calling & NeurIPS'23 & Not Available\\
\midrule

\shortstack{Reflexion: Language Agents with Verbal Reinforcement Learning$^{[2]}$} & Textual Feedback  & GPT-3 & API Calling & NeurIPS'23 & \href{https://github.com/noahshinn/reflexion}{Link}\\
\midrule

\shortstack{Iterative Translation Refinement with Large Language Models$^{[3]}$} & Textual Feedback  & GPT-3.5 & API Calling &Arxiv'23 & Not Available\\
\midrule

\shortstack{Leveraging GPT-4 for Automatic Translation Post-Editing$^{[4]}$} & Textual Feedback  & Multiple LLMs &  API Calling& EMNLP'23 & Not Available\\
\midrule

\shortstack{A New Benchmark and Reverse Validation Method for Passage-level Hallucination Detection$^{[5]}$} & Textual Feedback  & ChatGPT & API Calling & EMNLP'23 & \href{https://github.com/maybenotime/PHD}{Link}\\
\midrule

\shortstack{SELFCHECKGPT: Zero-Resource Black-Box Hallucination Detection for Generative Large Language Models$^{[6]}$} & Textual Feedback & Multiple LLMs & API Calling \& Model Inference & EMNLP'23 & \href{https://github.com/potsawee/selfcheckgpt}{Link}\\
\midrule

\shortstack{Improving Factuality and Reasoning in Language Models through Multiagent Debate$^{[7]}$} & Textual Feedback - Peer Review & Multiple LLMs & API Calling &  & \href{https://composable-models.github.io/llm_debate/}{Link}\\
\midrule

\shortstack{Towards Reasoning in Large Language Models via Multi-Agent Peer Review Collaboration$^{[8]}$} & Textual Feedback - Peer Review & Multiple LLMs & API Calling & Arxiv'23 & \href{https://github.com/HITsz-TMG/Multi-agent-peer-review}{Link}\\
\midrule

\shortstack{LM vs LM: Detecting Factual Errors via Cross Examination$^{[9]}$} & Textual Feedback - Peer Review  & Multiple LLMs & API Calling \& Model Inference & EMNLP'23 & Not Available\\
\midrule

\shortstack{Improving Language Model Negotiation with Self-Play and In-Context Learning from AI Feedback$^{[10]}$} & Textual Feedback - Peer Review & Multiple LLMs & API Calling & Arxiv'23 & \href{https://github.com/FranxYao/GPT-Bargaining}{Link}\\
\midrule

\shortstack{PRD: Peer Rank and Discussion Improve Large Language Model based Evaluations$^{[11]}$} & Textual Feedback - Peer Review & Multiple LLMs & \begin{tabular}[c]{@{}c@{}} API Calling,\\ \$0.14 per sample \end{tabular} & Arxiv'23 & \href{https://bcdnlp.github.io/PR_LLM_EVAL/}{Link}\\
\midrule

\shortstack{PRE: A Peer Review Based Large Language Model Evaluator$^{[12]}$} & Textual Feedback - Peer Review & Multiple LLMs & API Calling & Arxiv'24 & Not Available\\
\midrule

\shortstack{PiCO: Peer Review in LLMs based on the Consistency Optimization$^{[13]}$} & Textual Feedback - Peer Review & Multiple LLMs & API Calling \& Model Inference & Arxiv'24 & Not Available\\
\midrule

\shortstack{Learning from Mistakes via Cooperative Study Assistant for Large Language Models$^{[14]}$} & Textual Feedback - Mistake & Multiple LLMs & Model Inference & EMNLP'23 & \href{https://dqwang122.github.io/projects/SALAM/}{Link}\\
\midrule

\shortstack{Learning From Mistakes Makes LLM Better Reasoner$^{[15]}$} & Textual Feedback - Mistake & GPT-4 & API Calling & Arxiv'23 & \href{https://github.com/microsoft/LEMA}{Link}\\
\midrule

\shortstack{GAINING WISDOM FROM SETBACKS: ALIGNING LARGE LANGUAGE MODELS VIA MISTAKE ANALYSIS$^{[16]}$} & Textual Feedback - Mistake & Multiple LLMs & API Calling \& Modeling Inference & ICLR'24 & Not Available\\
\midrule

\shortstack{Can LLMs Learn from Previous Mistakes? Investigating LLMs’ Errors to Boost for Reasoning$^{[17]}$} & Textual Feedback - Mistake & Multiple LLMs & API Calling \& Modeling Inference & ACL'24 & \href{https://github.com/YookiTong/Learn-from-Mistakes-CotErrorSet}{Link}\\
\midrule

\rowcolor{Gray}
\multicolumn{6}{c}{Other Domain-specific Data}\\
\midrule

\shortstack{SODA: Million-scale Dialogue Distillation with Social Commonsense Contextualization$^{[18]}$} & Dialogue & GPT-3.5 & \begin{tabular}[c]{@{}c@{}} API Calling,\\ \$0.02 per dialogue \end{tabular} & EMNLP'23 & \href{https://github.com/skywalker023/sodaverse}{Link}\\
\midrule

\shortstack{Baize: An Open-Source Chat Model with Parameter-Efficient Tuning on Self-Chat Data$^{[19]}$} & Dialogue & Alpaca & Model Inference & EMNLP'23 & \href{https://github.com/project-baize/baize-chatbot}{Link}\\
\midrule

\shortstack{PLACES: Prompting Language Models for Social Conversation Synthesis$^{[20]}$} & Dialogue & Multiple LLMs & Model Inference & EACL'24 & Not Available\\
\midrule

\shortstack{CAMEL: Communicative Agents for ``Mind'' Exploration of Large Language Model Society$^{[21]}$} & Dialogue & ChatGPT & API Calling & NuerIPS'23 & \href{https://github.com/camel-ai/camel}{Link}\\
\midrule

\shortstack{AUGESC: Dialogue Augmentation with Large Language Models for Emotional Support Conversation$^{[22]}$} & Dialogue & GPT-J & Model Inference & ACL'23 & \href{https://github.com/thu-coai/AugESC}{Link}\\
\midrule

\shortstack{Weakly Supervised Data Augmentation Through Prompting for Dialogue Understanding$^{[23]}$} & Dialogue & GPT-J & Model Inference & NeurIPS'22 & Not Available\\
\midrule

\shortstack{Reflect, Not Reflex: Inference-Based Common Ground Improves Dialogue Response Quality$^{[24]}$} & Dialogue & GPT-3 & API Calling & EMNLP'22 & \href{https://inklab.usc.edu/Reflect/}{Link}\\
\midrule

\shortstack{ASDOT: Any-Shot Data-to-Text Generation with Pretrained Language Models$^{[25]}$} & Context & GPT-3 & \begin{tabular}[c]{@{}c@{}} API Calling,\\ \$23 in total \end{tabular} & EMNLP'22 & \href{https://github.com/szxiangjn/any-shot-data2text}{Link}\\
\midrule

\shortstack{Contextualization Distillation from Large Language Model for Knowledge Graph Completion$^{[26]}$} & Context & PaLM-2 & API Calling & EACL'24 & \href{https://github.com/David-Li0406/Contextulization-Distillation}{Link}\\
\midrule

\shortstack{Towards Ontology-Enhanced Representation Learning for Large Language Models$^{[27]}$} & Context & ChatGPT & API Calling & Arxiv'24 & \href{https://github.com/iqvianlp/llm-onto-infuse/}{Link}\\
\midrule

\shortstack{DALK: Dynamic Co-Augmentation of LLMs and KG to answer Alzheimer’s Disease Questions with Scientific Literature$^{[28]}$} & Graph & ChatGPT & API Calling & Arxiv'24 & \href{https://github.com/David-Li0406/DALK}{Link}\\
\midrule

\shortstack{Automated Construction of Theme-specific Knowledge Graphs$^{[29]}$} & Graph & GPT-4 & API Calling & Arxiv'24 & Not Available\\
\midrule

\shortstack{Large Language Models Can Learn Temporal Reasoning$^{[30]}$} & Graph & GPT-3.5 & API Calling & ACL'24 & \href{https://github.com/xiongsiheng/TG-LLM}{Link}\\
\midrule

\shortstack{Moving from Tabular Knowledge Graph Quality Assessment to RDF Triples Leveraging ChatGPT$^{[31]}$} & Graph & ChatGPT & API Calling & Arxiv'24 & \href{https://github.com/isislab-unisa/KGHeartbeat/tree/main}{Link}\\
\midrule

\shortstack{Language Models as Zero-Shot Planners: Extracting Actionable Knowledge for Embodied Agents$^{[32]}$} & Plan & GPT-3 & API Calling & ICML'22 & \href{https://wenlong.page/language-planner/}{Link}\\
\midrule

\shortstack{Do As I Can, Not As I Say: Grounding Language in Robotic Affordances$^{[33]}$} & Plan & Multiple LLMs & API Calling \& Model Inference & CoRL'21 & \href{https://say-can.github.io/}{Link}\\
\midrule

\shortstack{SayPlan: Grounding Large Language Models using 3D Scene Graphs for Scalable Robot Task Planning$^{[34]}$} & Plan & GPT-3.5 & API Calling & CoRL'23 & \href{https://sayplan.github.io/}{Link}\\
\midrule

\shortstack{PROGPROMPT: Generating Situated Robot Task Plans using Large Language Models$^{[35]}$} & Plan & GPT-3 & API Calling & ICRA'23 & \href{progprompt.github.io}{Link}\\
\midrule

\shortstack{Text2Motion: From Natural Language Instructions to Feasible Plans$^{[36]}$} & Plan & GPT-3.5 & API Calling &  Autonomous Robots'23 & \href{https://sites.google.com/stanford.edu/text2motion}{Link}\\
\midrule

\shortstack{GENSIM: GENERATING ROBOTIC SIMULATION TASKS VIA LARGE LANGUAGE MODELS$^{[37]}$} & Simulation Task & GPT-4 & API Calling & ICLR'24 & \href{https://liruiw.github.io/gensim}{Link}\\
\midrule

\shortstack{Scaling Up and Distilling Down: Language-Guided Robot Skill Acquisition$^{[38]}$} & Simulation Task & Multiple LLMs & API Calling & CoRL'23 & \href{https://www.cs.columbia.edu/ huy/scalingup/}{Link}\\
\midrule

\shortstack{REWARD DESIGN WITH LANGUAGE MODELS$^{[39]}$} & Reward & GPT-3 & API Calling & ICLR'23 & \href{https://github.com/minaek/reward_design_with_llms}{Link}\\
\midrule

\shortstack{Guiding Pretraining in Reinforcement Learning with Large Language Models$^{[40]}$} & Reward & GPT-3 & \begin{tabular}[c]{@{}c@{}} API Calling,\\ 0.02 second per call \end{tabular} & ICML'23 & Not Available\\
\midrule

\shortstack{Enhanced Visual Instruction Tuning with Synthesized Image-Dialogue Data$^{[41]}$} & Visual Instruction Tuning Data & ChatGPT & API Calling & Arxiv'23 & \href{https://github.com/icoz69/StableLLAVA}{Link}\\
\midrule

\shortstack{LAMM: Language-Assisted Multi-Modal Instruction-Tuning Dataset, Framework, and Benchmark$^{[42]}$} & Visual Instruction Tuning Data & GPT-4 & API Calling & NeurIPS'23 & \href{https://openlamm.github.io}{Link}\\
\midrule

\shortstack{TOMGPT: Reliable Text-Only Training Approach for Cost-Efective Multi-modal Large Language Model$^{[43]}$} & Context & ChatGPT & API Calling & TKDD'24 & Not Available\\
\midrule

\shortstack{LLM Based Generation of Item-Description for Recommendation System$^{[44]}$} & Item Description &  Alpaca & Model Inference & RecSys'23 & Not Available\\
\midrule

\shortstack{PMG : Personalized Multimodal Generation with Large Language$^{[45]}$} & Context & Multiple LLMs & Model Inference & WWW'24 & \href{https://github.com/mindspore-lab/models/tree/master/research/huawei-noah/PMG}{Link}\\
\midrule

\shortstack{LLMRec: Large Language Models with Graph Augmentation for Recommendation$^{[46]}$} & Augmented Implicit
Feedback  & ChatGPT & API Calling, \$21.14 & WSDM'24  & \href{https://github.com/HKUDS/LLMRec.git}{Link}\\
\midrule

\shortstack{Large Language Models as Evaluators for Recommendation Explanations$^{[47]}$} & Explanation & Multiple LLMs & API Calling \& Model Inference, less than \$0.02 per sample & Arxiv'24 & \href{https://github.com/Xiaoyu-SZ/LLMasEvaluator}{Link}\\
\midrule

\shortstack{Exploiting Asymmetry for Synthetic Training Data Generation: SynthIE and the Case of Information Extraction$^{[48]}$} & IE Sample & GPT-3.5 & API Calling, \$223.55 for entire dataset & EMNLP'23 & \href{https://github.com/epfl-dlab/SynthIE}{Link}\\
\midrule

\shortstack{InPars-v2: Large Language Models as Efficient Dataset Generators for Information Retrieval$^{[49]}$} & IE sample & GPT-J & \begin{tabular}[c]{@{}c@{}} Model Inference,\\ 30 hours on an A100 GPU to generate 100k queries \end{tabular} & Arxiv'23 & \href{https://github.com/zetaalphavector/inPars/tree/master/tpu}{Link}\\
\midrule

\shortstack{READ: Improving Relation Extraction from an ADversarial Perspective$^{[50]}$} & IE Sample & ChatGPT & API Calling & NAACL'24 & \href{https://github.com/David-Li0406/READ}{Link}\\
\midrule

\shortstack{STAR: Boosting Low-Resource Information Extraction by Structure-to-Text Data Generation with Large Language Models$^{[51]}$} & IE Sample & Multiple LLMs & API Calling & AAAI'24 & \href{https://derek.ma/STAR}{Link}\\
\midrule

\shortstack{Adjudicating LLMs as PropBank Annotators$^{[52]}$} & IE Label & Multiple LLMs & API Calling & LREC'24 & \href{https://github.com/H-TayyarMadabushi/Adjudicating-LLMs-as-PropBank-Annotators}{Link}\\
\midrule

\shortstack{A Causal Explainable Guardrails for Large Language Models$^{[53]}$} & Representation & GPT-4 & API Calling & Arxiv'24  & Not Available\\
\midrule

\shortstack{Zero-shot LLM-guided Counterfactual Generation for Text$^{[54]}$} & Context & Multiple LLMs & API Calling & Arxiv'24 & Not Available\\
\midrule

\shortstack{Text classification of column headers with a controlled vocabulary: leveraging LLMs for metadata enrichment$^{[55]}$} & Metadata & ChatGPT & API Calling & Arxiv'24 & \href{https://github.com/ritamargherita/LLMs-topic-classification/}{Link}\\
\midrule

\toprule
\end{tabular}
}
\begin{tablenotes}
\item 
\small{Note: $^{[1]}$\cite{madaan2024self};
$^{[2]}$\cite{shinn2024reflexion};
$^{[3]}$\cite{chen2023iterative};
$^{[4]}$\cite{raunak2023leveraging};
$^{[5]}$\cite{yang2023new};
$^{[6]}$\cite{manakul2023selfcheckgpt};
$^{[7]}$\cite{du2023improving};
$^{[8]}$\cite{xu2023towards};
$^{[9]}$\cite{cohen2023lm};
$^{[10]}$\cite{fu2023improving};
$^{[11]}$\cite{li2023prd};
$^{[12]}$\cite{chu2024pre};
$^{[13]}$\cite{ning2024peer};
$^{[14]}$\cite{wang2023learning};
$^{[15]}$\cite{an2023learning};
$^{[16]}$\cite{chen2023gaining};
$^{[17]}$\cite{tong2024can};
$^{[18]}$\cite{kim2023soda};
$^{[19]}$\cite{xu2023baize};
$^{[20]}$\cite{chen2023places};
$^{[21]}$\cite{li2024camel};
$^{[22]}$\cite{zheng2023augesc};
$^{[23]}$\cite{chen2022weakly};
$^{[24]}$\cite{zhou2022reflect};
$^{[25]}$\cite{xiang2022asdot};
$^{[26]}$\cite{li2024contextualization};
$^{[27]}$\cite{ronzano2024towards};
$^{[28]}$\cite{li2024dalk};
$^{[29]}$\cite{ding2024automated};
$^{[30]}$\cite{xiong2024large};
$^{[31]}$\cite{tuozzo2022moving};
$^{[32]}$\cite{huang2022language};
$^{[33]}$\cite{brohan2023can};
$^{[34]}$\cite{rana2023sayplan};
$^{[35]}$\cite{singh2023progprompt};
$^{[36]}$\cite{lin2023text2motion};
$^{[37]}$\cite{wang2023gensim};
$^{[38]}$\cite{ha2023scaling};
$^{[39]}$\cite{kwon2022reward};
$^{[40]}$\cite{du2023guiding};
$^{[41]}$\cite{li2023stablellava};
$^{[42]}$\cite{yin2024lamm};
$^{[43]}$\cite{chen2024tomgpt};
$^{[44]}$\cite{acharya2023llm};
$^{[45]}$\cite{shen2024pmg};
$^{[46]}$\cite{wei2024llmrec};
$^{[47]}$\cite{zhang2024large};
$^{[48]}$\cite{josifoski2023exploiting};
$^{[49]}$\cite{jeronymo2023inpars};
$^{[50]}$\cite{li2024read};
$^{[51]}$\cite{ma2024star};
$^{[52]}$\cite{bonn2024adjudicating};
$^{[53}$\cite{chu2024causal};
$^{[54]}$\cite{bhattacharjee2024zero};
$^{[55]}$\cite{martorana2024text}.}
\end{tablenotes}
\caption{A list of representative LLM-Based Annotation Generation (Textual Feedback, Other Domain-specific Data) papers with open-source code/data.} 
\label{tab:LLM-Based Annotation Generation-2}
\end{table*}
%%%%%%%%%%%%%%%%%%%%%%%%%%%%%%%%%%%%%%%%%%%%%%%%%%%%%%%%%%%%%%%

\begin{table*}[t]
\centering
% \scriptsize
\resizebox{1\textwidth}{!}{
\begin{tabular} {c|c|c|c|c|c}
\rowcolor{Gray}
\toprule
Paper & Data Type & Backbone & Annotation Cost & Venue & Code/Data Link \\
\midrule

\rowcolor{Gray}
\multicolumn{6}{c}{Filter \& Selection}\\
\midrule

\shortstack{Constitutional AI: Harmlessness from AI Feedback$^{[1]}$} & Pairwise Feedback & Multiple LLMs & Model Inference & Arxiv'22 & \href{https://github.com/anthropics/ConstitutionalHarmlessnessPaper}{Link}\\
\midrule

\shortstack{SODA: Million-scale Dialogue Distillation with Social Commonsense Contextualization$^{[2]}$} & Dialogue & GPT-3.5 & \begin{tabular}[c]{@{}c@{}} API Calling,\\ \$0.02 per dialogue \end{tabular} & EMNLP'23 & \href{https://github.com/skywalker023/sodaverse}{Link}\\
\midrule

\shortstack{Aligning Large Language Models through Synthetic Feedback$^{[3]}$} & Pairwise Feedback & LLaMA & Model Inference & EMNLP'23 & \href{https://github.com/naver-ai/almost}{Link}\\
\midrule

\shortstack{AUGESC: Dialogue Augmentation with Large Language Models for Emotional Support Conversation$^{[4]}$} & Dialogue & GPT-J & Model Inference & ACL'23 & \href{https://github.com/thu-coai/AugESC}{Link}\\
\midrule

\shortstack{SELF-QA: Unsupervised Knowledge Guided Language Model Alignment$^{[5]}$} & Instruction \& Response &  BLOOM & Model Inference & Arxiv'23 & Not Available \\
\midrule

\shortstack{Human-Instruction-Free LLM Self-Alignment with Limited Samples$^{[6]}$} & Instruction \& Response & Multiple LLMs & \begin{tabular}[c]{@{}c@{}}  Model Inference,\\ single NVIDIA A100 80G GPU   \end{tabular} & Arxiv'24 & Not Available\\
\midrule

\shortstack{Automated Construction of Theme-specific Knowledge Graphs$^{[7]}$} & Graph & GPT-4 & API Calling & Arxiv'24 & Not Available\\
\midrule

\shortstack{Large Language Models Are Reasoning Teachers$^{[8]}$} & CoT & GPT-3.5 & API Calling & ACL'23 & \href{https://github.com/itsnamgyu/reasoning-teacher}{Link}\\
\midrule

\shortstack{Knowledge-Augmented Reasoning Distillation for Small Language Models in Knowledge-Intensive Tasks
$^{[9]}$} & Rationale - CoT & ChatGPT & API Calling & NeurIPS'23 & \href{https://github.com/Nardien/KARD}{Link}\\
\midrule

\shortstack{SELF-CONSISTENCY IMPROVES CHAIN OF THOUGHT REASONING IN LANGUAGE MODELS$^{[10]}$} & Rationale - Diverse Thinking & Multiple LLMs & API Calling \& Model Inference & ICLR'23 & Not Available\\
\midrule

\shortstack{Making Large Language Models Better Data Creators$^{[11]}$} & Instruction \& Response & ChatGPT & API Calling & EMNLP'23 & \href{https://github.com/microsoft/llm-data-creation}{Link}\\
\midrule

\shortstack{Automated Construction of Theme-specific Knowledge Graphs$^{[12]}$} & Graph & GPT-4 & API Calling & Arxiv'24 & Not Available\\
\midrule

\shortstack{Reinforced Self-Training (ReST) for Language Modeling$^{[13]}$} & Response & Multiple LLMs & Model Inference & Arxiv'24 & Not Available\\
\midrule

\shortstack{RAFT: Reward rAnked FineTuning for Generative Foundation Model Alignment$^{[14]}$} & Response & LLaMA & Model Inference & TMLR & \href{ https://github.com/OptimalScale/LMFlow}{Link}\\
\midrule

\shortstack{Selective In-Context Data Augmentation for Intent Detection using Pointwise V-Information$^{[15]}$} & Instruction & OPT & Model Inference & EACL'24 & Not Available\\
\midrule

\shortstack{CodecLM: Aligning Language Models with Tailored Synthetic Data$^{[16]}$} & Instruction & LLaMA & Model Inference & NAACL'24 & Not Available\\
\midrule

\shortstack{DISCO: Distilling Counterfactuals with Large Language Models$^{[17]}$} & CoT & GPT-3 & API Callin & ACL'23 & \href{https://github.com/eric11eca/disco}{Link}\\
\midrule

\shortstack{LARGE LANGUAGE MODELS CAN SELF-IMPROVE$^{[18]}$} & Response &  PaLM-540B & Model Inference & EMNLP'23 & Not Available \\
\midrule

\shortstack{West-of-N: Synthetic Preference Generation for Improved Reward Modeling$^{[19]}$} & Pairwise Feedback &  T5-XXL & Model Inference  & Arxiv'24 & Not Available\\
\midrule

\shortstack{SELF: SELF-EVOLUTION WITH LANGUAGE FEEDBACK$^{[20]}$} & Response & Multiple LLMs & Model Inference & Arxiv'23 & Not Available\\
\midrule

\shortstack{InPars-v2: Large Language Models as Efficient Dataset Generators for Information Retrieval$^{[21]}$} & IE sample & GPT-J & \begin{tabular}[c]{@{}c@{}} Model Inference,\\ 30 hours on an A100 GPU to generate 100k queries \end{tabular} & Arxiv'23 & \href{https://github.com/zetaalphavector/inPars/tree/master/tpu}{Link}\\
\midrule

\shortstack{DALK: Dynamic Co-Augmentation of LLMs and KG to answer Alzheimer’s Disease Questions with Scientific Literature$^{[22]}$} & Graph & ChatGPT & API Calling & Arxiv'24 & \href{https://github.com/David-Li0406/DALK}{Link}\\
\midrule

\shortstack{Optimizing Language Model’s Reasoning Abilities with Weak Supervision$^{[23]}$} & Pairwise Feedback &  LLaMA & Model Inference & Arxiv'24 & Not Available\\
\midrule

\toprule
\end{tabular}
}
\begin{tablenotes}
\item 
\small{Note: $^{[1]}$\cite{bai2022constitutional};
$^{[2]}$\cite{kim2023soda};
$^{[3]}$\cite{kim2023aligning};
$^{[4]}$\cite{zheng2023augesc};
$^{[5]}$\cite{zhang2023self};
$^{[6]}$\cite{guo2024human};
$^{[7]}$\cite{ding2024automated};
$^{[8]}$\cite{ho2023large};
$^{[9]}$\cite{kang2024knowledge};
$^{[10]}$\cite{wang2022self};
$^{[11]}$\cite{lee2023making};
$^{[12]}$\cite{ding2024automated};
$^{[13]}$\cite{gulcehre2023reinforced};
$^{[14]}$\cite{dong2023raft};
$^{[15]}$\cite{lin2023selective};
$^{[16]}$\cite{wang2024codeclm};
$^{[17]}$\cite{chen2023disco};
$^{[18]}$\cite{huang2023large};
$^{[19]}$\cite{pace2024west};
$^{[20]}$\cite{lu2023self};
$^{[21]}$\cite{jeronymo2023inpars};
$^{[22]}$\cite{li2024dalk};
$^{[23]}$\cite{tong2024optimizing}.}
\end{tablenotes}
\caption{A list of representative LLM-Generated Annotation Assessment papers with open-source code/data.} 
\label{tab:LLM-Generated Annotation Assessment}
\end{table*}
%%%%%%%%%%%%%%%%%%%%%%%%%%%%%%%%%%%%%%%%%%%%%%%%%%%%%%%%%%%%%%%

\begin{table*}[t]
\centering
% \scriptsize
\resizebox{1\textwidth}{!}{
\begin{tabular} {c|c|c|c|c|c}
\rowcolor{Gray}
\toprule
Paper & Data Type & Backbone & Annotation Cost & Venue & Code/Data Link \\
\midrule

\rowcolor{Gray}
\multicolumn{6}{c}{Supervised Fine-tuning}\\
\midrule

\shortstack{LARGE LANGUAGE MODELS CAN SELF-IMPROVE$^{[1]}$} & Response &  PaLM-540B & Model Inference & EMNLP'23 & Not Available \\
\midrule

\shortstack{SELF-INSTRUCT: Aligning Language Models with Self-Generated Instructions$^{[2]}$} & Instruction \& Response & GPT-3 & \begin{tabular}[c]{@{}c@{}} API Calling,\\ \$600 for entire dataset \end{tabular} & ACL'23 & \href{https://github.com/yizhongw/self-instruct}{Link}\\
\midrule

\shortstack{SELF: SELF-EVOLUTION WITH LANGUAGE FEEDBACK$^{[3]}$} & Response & Multiple LLMs & Model Inference & Arxiv'23 & Not Available\\
\midrule

\shortstack{Self-Distillation Bridges Distribution Gap in Language Model Fine-Tuning$^{[4]}$} & Response & LLaMA-2 & Model Inference & ACL'24 & \href{https://github.com/sail-sg/sdft}{Link}\\
\midrule

\shortstack{Self-Play Fine-Tuning Converts Weak Language Models to Strong Language Models$^{[5]}$} & Response  & zephyr & Model Inference & Arxiv'24 & \href{https://github.com/uclaml/SPIN}{Link}\\
\midrule

\shortstack{Self-playing Adversarial Language Game Enhances LLM Reasoning$^{[6]}$} & Response  & Multiple LLMs & Model Inference & Arxiv'24 & \href{https://github.com/Linear95/SPAG}{Link}\\
\midrule

\shortstack{Stanford alpaca: An instruction-following llama model$^{[7]}$} &   Response & GPT-3.5 & API Calling & Arxiv'23 & \href{https://github.com/tatsu-lab/stanford_alpaca}{Link}\\
\midrule

\shortstack{Vicuna: An open-source chatbot impressing gpt-4 with 90\%* chatgpt quality$^{[8]}$} &  Response & GPT-4 & API Calling & Arxiv'23 & \href{https://lmsys.org/blog/2023-03-30-vicuna/}{Link}\\
\midrule

\shortstack{Wizardlm: Empowering large language models to follow complex instructions$^{[9]}$} & Instruction  & LLaMA & Model Inference & Arxiv'23 & \href{https://github.com/nlpxucan/WizardLM}{Link}\\
\midrule

\shortstack{Generating training data with language models: Towards zero-shot language understanding$^{[10]}$} &  Instruction & CTRL & Model Inference & NeurIPS & \href{https://github.com/yumeng5/SuperGen}{Link}\\
\midrule

\shortstack{Tuning Language Models as Training Data Generators for Augmentation-Enhanced Few-Shot Learning$^{[11]}$} & Instruction &  CTRL & Model Training  & ICML'23 & \href{https://github.com/yumeng5/FewGen}{Link}\\
\midrule

\shortstack{Noise-Robust Fine-Tuning of Pretrained Language Models via External Guidance$^{[12]}$} & Response  & ChatGPT & API Calling & EMNLP'23 & \href{https://github.com/SongW-SW/LAFT}{Link}\\
\midrule

\shortstack{PINTO: FAITHFUL LANGUAGE REASONING USING PROMPT-GENERATED RATIONALES$^{[13]}$} & Rationale - CoT & GPT-neox & Model Inference & ICLR'23 & \href{https://github.com/wangpf3/pinto-faithful-language-reasoning}{Link}\\
\midrule

\shortstack{Distilling Reasoning Capabilities into Smaller Language Models$^{[14]}$} & Rationale - CoT & GPT-3 & API Calling & ACL'23 & Not Available\\
\midrule

\shortstack{LogiCoT: Logical Chain-of-Thought Instruction Tuning$^{[15]}$} & Rationale - CoT & GPT-4 & API Calling & EMNLP'23 & Not Available \\
\midrule

\shortstack{Knowledge-Augmented Reasoning Distillation for Small Language Models in Knowledge-Intensive Tasks$^{[16]}$} & Rationale - CoT & ChatGPT & API Calling & NeurIPS'23 & \href{https://github.com/Nardien/KARD}{Link}\\
\midrule

\shortstack{Baize: An Open-Source Chat Model with Parameter-Efficient Tuning on Self-Chat Data$^{[17]}$} & Dialogue & Alpaca & Model Inference & EMNLP'23 & \href{https://github.com/project-baize/baize-chatbot}{Link}\\
\midrule

\shortstack{Exploiting Asymmetry for Synthetic Training Data Generation: SynthIE and the Case of Information Extraction$^{[18]}$} & IE Sample & GPT-3.5 & API Calling, \$223.55 for entire dataset & EMNLP'23 & \href{https://github.com/epfl-dlab/SynthIE}{Link}\\
\midrule

\shortstack{InPars-v2: Large Language Models as Efficient Dataset Generators for Information Retrieval$^{[19]}$} & IE sample & GPT-J & \begin{tabular}[c]{@{}c@{}} Model Inference,\\ 30 hours on an A100 GPU to generate 100k queries \end{tabular} & Arxiv'23 & \href{https://github.com/zetaalphavector/inPars/tree/master/tpu}{Link}\\
\midrule

\shortstack{Code alpaca: An instruction-following llama model for code generation$^{[20]}$} &  Instruction \& Response & Alpaca & Model Inferece & Arxiv'23 & \href{https://github.com/sahil280114/codealpaca}{Link}\\
\midrule

\shortstack{Code llama: Open foundation models for code$^{[21]}$} & Instruction \& Response  & Multiple LLMs & Model Inference & Arxiv'23 & \href{https://github.com/facebookresearch/codellama}{Link}\\
\midrule

\shortstack{HuatuoGPT, Towards Taming Language Model to Be a Doctor$^{[22]}$} & Instruction \& Response & ChatGPT & API Calling & Arxiv'23 & \href{https://github.com/FreedomIntelligence/HuatuoGPT}{Link}\\
\midrule

\shortstack{Doctorglm: Fine-tuning your chinese doctor is not a herculean task$^{[23]}$} & Response  & ChatGPT & API Calling & Arxiv'23 & \href{https://github.com/xionghonglin/DoctorGLM}{Link}\\
\midrule

\shortstack{Xuanyuan 2.0: A large chinese financial chat model with hundreds of billions parameters$^{[24]}$} &  Instruction \& Response & BLOOM & Model Inference & CIKM'23 & Not Available\\
\midrule

\shortstack{Wizardmath: Empowering mathematical reasoning for large language models via reinforced evol-instruct$^{[25]}$} & Pairwise Feedback & ChatGPT & API Calling & Arxiv'23 & \href{https://github.com/nlpxucan/WizardLM}{Link}\\
\midrule

\shortstack{Gimlet: A unified graph-text model for instruction-based molecule zero-shot learning$^{[26]}$} & Instruction  & ChatGPT & API Calling & NuerIPS'23 & \href{https://github.com/zhao-ht/GIMLET}{Link}\\
\midrule

\rowcolor{Gray}
\multicolumn{6}{c}{Alignment Tuning}\\
\midrule

\shortstack{Automatic Pair Construction for Contrastive Post-training$^{[27]}$} & Pairwise Feedback & LLaMA & \begin{tabular}[c]{@{}c@{}}  Model Inference,\\ 16 Nvidia V100 GPUs   \end{tabular}& NAACL'24 & Not Available\\
\midrule

\shortstack{Aligning Large Language Models through Synthetic Feedback$^{[28]}$} & Pairwise Feedback & LLaMA & Model Inference & EMNLP'23 & \href{https://github.com/naver-ai/almost}{Link}\\
\midrule

\shortstack{West-of-N: Synthetic Preference Generation for Improved Reward Modeling$^{[29]}$} & Pairwise Feedback &  T5-XXL & Model Inference  & Arxiv'24 & Not Available\\
\midrule

\shortstack{Learning Reward for Robot Skills Using Large Language Models via Self-Alignment$^{[30]}$} & Pairwise Feedback & ChatGPT & API Calling & ICML'24 & \href{https://sites.google.com/view/rewardselfalign}{Link}\\
\midrule

\shortstack{SALMON: SELF-ALIGNMENT WITH INSTRUCTABLE REWARD MODELS$^{[31]}$} & Pairwise Feedback & LLaMA-2 & Model Inference & ICLR'24 & \href{https://github.com/IBM/SALMON}{Link}\\
\midrule

\shortstack{Self-Rewarding Language Models$^{[32]}$} & Pairwise Feedback & LLaMA-2 & Model Inference & Arxiv'24 &  Not Available \\
\midrule

\shortstack{Self-Alignment for Factuality: Mitigating Hallucinations in LLMs via Self-Evaluation$^{[33]}$} & Pairwise Feedback & LLaMA & Model Inference & Arxiv'24 & \href{https://github.com/zhangxy-2019/Self-Alignment-for-Factuality}{Link}\\
\midrule

\shortstack{Aligning Large Language Models by On-Policy Self-Judgment$^{[34]}$} &  Response & LLaMA-2 & Model Inference & Arxiv'24 & \href{https://github.com/oddqueue/self-judge}{Link}\\
\midrule

\shortstack{Optimizing Language Model’s Reasoning Abilities with Weak Supervision$^{[35]}$} & Pairwise Feedback &  LLaMA & Model Inference & Arxiv'24 & Not Available\\
\midrule

\shortstack{Reinforcement Learning from Reflective Feedback (RLRF): Aligning and Improving LLMs via Fine-Grained Self-Reflection$^{[36]}$} & Pairwise Feedback & LLaMA-2 & \begin{tabular}[c]{@{}c@{}}  Model Inference,\\ 16 Nvidia V100 GPUs   \end{tabular} & Arxiv'24 & Not Available\\
\midrule

\shortstack{Direct language model alignment from online ai feedback$^{[37]}$} & Pairwise Feedback  & PaLM-2 & API Calling & Arxiv'24 & Not Available\\
\midrule

\shortstack{Reinforced Self-Training (ReST) for Language Modeling$^{[38]}$} & Response & Multiple LLMs & Model Inference & Arxiv'24 & Not Available\\
\midrule

\shortstack{RAFT: Reward rAnked FineTuning for Generative Foundation Model Alignment$^{[39]}$} & Response & LLaMA & Model Inference & TMLR & \href{ https://github.com/OptimalScale/LMFlow}{Link}\\
\midrule

\shortstack{Step-On-Feet Tuning: Scaling Self-Alignment of LLMs via Bootstrapping$^{[40]}$} & Response & LLaMA-2 & Model Inference & Arxiv'24 & Not Available\\
\midrule

\shortstack{Mixture of insighTful Experts (MoTE): The Synergy of Thought Chains and Expert Mixtures in Self-Alignment$^{[41]}$} & Response & Alpaca & Model Inference & Arxiv'24 & Not Available\\
\midrule

\shortstack{Iterative reasoning preference optimization$^{[42]}$} & Pairwise Feedback & LLaMA-2 & Model Inference & Arxiv'24 & Not Available\\
\midrule

\rowcolor{Gray}
\multicolumn{6}{c}{Inference Time}\\
\midrule

\shortstack{Large Language Models are Human-Level Prompt Engineers$^{[43]}$} & Instruction  & GPT-3.5 & API Calling & ICLR'23 & \href{https://github.com/keirp/automatic_prompt_engineer}{Link}\\
\midrule

\shortstack{Auto-ICL: In-Context Learning without Human Supervision$^{[44]}$} &  Instruction & ChatGPT & API Calling & Arxiv'23 & \href{https://github.com/ecielyang/Auto-ICL}{Link}\\
\midrule

\shortstack{Empowering Large Language Models for Textual Data Augmentation$^{[45]}$} &  Instruction & ChatGPT & API Calling & Arxiv'24 & Not Available\\
\midrule

\shortstack{Self-generated in-context learning: Leveraging auto-regressive language models as a demonstration generator$^{[46]}$} &  Instruction & GPT-J & Model Inference  & NAACL'22 & \href{}{Link}\\
\midrule

\shortstack{Are Human-generated Demonstrations Necessary for In-context Learning?$^{[47]}$} & Instruction  & Multiple LLMs & API Calling & Arxiv'23 & \href{https://github.com/ruili33/SEC}{Link}\\
\midrule

\shortstack{Self-ICL: Zero-Shot In-Context Learning with Self-Generated Demonstrations$^{[48]}$} & Instruction  & Multiple LLMs & API Calling & EMNLP'23 & \href{https://github.com/ntunlplab/Self-ICL}{Link}\\
\midrule

\shortstack{Self-Demos: Eliciting Out-of-Demonstration Generalizability in Large Language Models$^{[49]}$} & Instruction  & ChatGPT & API Calling & NAACL'24 & \href{https://github.com/hewei2001/Self-Demos}{Link}\\
\midrule

\shortstack{Rephrase and respond: Let large language models ask better questions for themselves$^{[50]}$} & Instruction  & GPT-4 & API Calling & Ariv'23 & \href{https://github.com/uclaml/Rephrase-and-Respond}{Link}\\
\midrule

\shortstack{DAIL: Data Augmentation for In-Context Learning via Self-Paraphrase$^{[51]}$} & Instruction & ChatGPT & API Calling & Arxiv'23 & Not Available \\
\midrule

\shortstack{Just rephrase it! Uncertainty estimation in closed-source language models via multiple rephrased queries$^{[52]}$} & Instruction  & Multiple LLMs & Model Inference & Arxiv'24 & Not Available\\
\midrule

\shortstack{Self-Polish: Enhance Reasoning in Large Language Models via Problem Refinement$^{[53]}$} &  Instruction & GPT-3.5 & API Calling & EMNLP'23 & \href{https://github.com/WooooDyy/Self-Polish}{Link}\\
\midrule

\shortstack{Self-DC: When to retrieve and When to generate? Self Divide-and-Conquer for Compositional Unknown Questions$^{[54]}$} & Instruction  & ChatGPT & API Calling & Arxiv'24 & Not Available\\
\midrule

\shortstack{Large Language Models are Zero-Shot Reasoners$^{[55]}$} & Rationale - CoT & Multiple LLMs & API Callinfg & NeurIPS'22 & Not Available \\
\midrule

\shortstack{SELF-CONSISTENCY IMPROVES CHAIN OF THOUGHT REASONING IN LANGUAGE MODELS$^{[56]}$} & Rationale - Diverse Thinking & Multiple LLMs & API Calling \& Model Inference & ICLR'23 & Not Available\\
\midrule

\shortstack{UNIVERSAL SELF-CONSISTENCY FOR LARGE LANGUAGE MODEL GENERATION$^{[57]}$} & Rationale - Diverse Thinking & Multiple LLMs & API Calling & Arxiv'23 & Not Available \\
\midrule

\shortstack{Eliminating Reasoning via Inferring with Planning: A New Framework to Guide LLMs’ Non-linear Thinking$^{[58]}$} & Rationale - Elimination & PaLM2 & API Calling & Arxiv'23 & Not Available\\
\midrule

\shortstack{It’s Not Easy Being Wrong: Large Language Models Struggle with Process of Elimination Reasoning$^{[59]}$} & Rationale - Elimination & Multiple LLMs & API Calling & ACL'24 & \href{https://github.com/nbalepur/PoE}{Link}\\
\midrule

\shortstack{POE: Process of Elimination for Multiple Choice Reasoning$^{[60]}$} & Rationale - Elimination & FLAN-T5 & Model Inference & EMNLP'23 & \href{https://github.com/KasMasVan/PoE}{Link}\\
\midrule

\shortstack{SELF-REFINE: Iterative Refinement with Self-Feedback$^{[61]}$} & Textual Feedback & Multiple LLMs & API Calling & NeurIPS'23 & Not Available\\
\midrule

\shortstack{Can LLMs Learn from Previous Mistakes? Investigating LLMs’ Errors to Boost for Reasoning$^{[62]}$} & Textual Feedback - Mistake & Multiple LLMs & API Calling \& Modeling Inference & ACL'24 & \href{https://github.com/YookiTong/Learn-from-Mistakes-CotErrorSet}{Link}\\
\midrule

\shortstack{Program of Thoughts Prompting: Disentangling Computation from Reasoning for Numerical Reasoning Tasks
$^{[63]}$} & Rationale - Program & Multiple LLMs & API Calling \& Model Inference & TMLR'23 & Not Available\\
\midrule

\shortstack{Graph of Thoughts: Solving Elaborate Problems with Large Language Models$^{[64]}$} & Rationale - Graph & GPT-3.5 & API Calling & AAAI'24 & \href{https://github.com/spcl/graph-of-thoughts}{Link}\\
\midrule

\shortstack{Reasoning with Language Model is Planning with World Model$^{[65]}$} & Rationale - Tree &  LLaMA & \begin{tabular}[c]{@{}c@{}}  Model Inference,\\ 4×24 GB NVIDIA A5000 GPUs   \end{tabular} & EMNLP'23 & \href{https://github.com/maitrix-org/llm-reasoners}{Link}\\
\midrule

\toprule
\end{tabular}
}
\begin{tablenotes}
\item 
\small{Note: $^{[1]}$\cite{huang2023large};
$^{[2]}$\cite{wang2023self};
$^{[3]}$\cite{lu2023self};
$^{[4]}$\cite{yang2024self};
$^{[5]}$\cite{chen2024self};
$^{[6]}$\cite{cheng2024self};
$^{[7]}$\cite{taori2023stanford};
$^{[8]}$\cite{vicuna2023};
$^{[9]}$\cite{xu2023wizardlm};
$^{[10]}$\cite{meng2022generating};
$^{[11]}$\cite{meng2023tuning};
$^{[12]}$\cite{wang2023noise};
$^{[13]}$\cite{wang2022pinto};
$^{[14]}$\cite{shridhar2023distilling};
$^{[15]}$\cite{liu2023logicot};
$^{[16]}$\cite{kang2024knowledge};
$^{[17]}$\cite{xu2023baize};
$^{[18]}$\cite{josifoski2023exploiting};
$^{[19]}$\cite{jeronymo2023inpars};
$^{[20]}$\cite{chaudhary2023code};
$^{[21]}$\cite{roziere2023code};
$^{[22]}$\cite{zhang2023huatuogpt};
$^{[23]}$\cite{xiong2023doctorglm};
$^{[24]}$\cite{zhang2023xuanyuan};
$^{[25]}$\cite{luo2023wizardmath};
$^{[26]}$\cite{zhao2024gimlet};
$^{[27]}$\cite{xu2023contrastive};
$^{[28]}$\cite{kim2023aligning};
$^{[29]}$\cite{pace2024west};
$^{[30]}$\cite{zeng2024learning};
$^{[31]}$\cite{Sun2023SALMONSW};
$^{[32]}$\cite{yuan2024self};
$^{[33]}$\cite{zhang2024self};
$^{[34]}$\cite{lee2024aligning};
$^{[35]}$\cite{tong2024optimizing};
$^{[36]}$\cite{lee2024reinforcement};
$^{[37]}$\cite{guo2024direct};
$^{[38]}$\cite{gulcehre2023reinforced};
$^{[39]}$\cite{dong2023raft};
$^{[40]}$\cite{wang2024step};
$^{[41]}$\cite{liu2024mixture};
$^{[42]}$\cite{chen2023iterative};
$^{[43]}$\cite{zhou2022large};
$^{[44]}$\cite{yang2023auto};
$^{[45]}$\cite{liempowering};
$^{[46]}$\cite{kim2022self};
$^{[47]}$\cite{li2023human};
$^{[48]}$\cite{chen2023self};
$^{[49]}$\cite{he2024self};
$^{[50]}$\cite{deng2023rephrase};
$^{[51]}$\cite{li2023dail};
$^{[52]}$\cite{yang2024just};
$^{[53]}$\cite{xi2023self};
$^{[54]}$\cite{wang2024self};
$^{[55]}$\cite{kojima2022large};
$^{[56]}$\cite{wang2022self};
$^{[57]}$\cite{chen2023universal};
$^{[58]}$\cite{tong2023eliminating};
$^{[59]}$\cite{balepur2023s};
$^{[60]}$\cite{ma2023poe};
$^{[61]}$\cite{madaan2024self};
$^{[62]}$\cite{tong2024can};
$^{[63]}$\cite{chen2023program};
$^{[64]}$\cite{besta2024graph};
$^{[65]}$\cite{hao2023reasoning}.}
\end{tablenotes}
\caption{A list of representative LLM-Generated Annotation Utilization papers with open-source code/data.} 
\label{tab:LLM-Generated Annotation Utilization}
\end{table*}

\end{document}